\documentclass[conference]{IEEEtran}
\usepackage{times}

\usepackage[numbers]{natbib}
\usepackage{multicol}
\usepackage[bookmarks=true]{hyperref}

\usepackage{amsfonts,amsmath}
\usepackage{bbm}
\usepackage{graphicx}
\usepackage{makecell}
\usepackage{xcolor}
\usepackage[capitalise]{cleveref}
\usepackage{setspace}
\usepackage{wrapfig,subcaption}
\usepackage{booktabs}

\newcommand{\rev}[1]{\color{black}{#1}\color{black}}
\usepackage[normalem]{ulem}

\definecolor{ForestGreen}{RGB}{34,139,34}
\usepackage{siunitx}

\begin{document}

\title{Can We Detect Failures Without Failure Data? Uncertainty-Aware Runtime Failure Detection for Imitation Learning Policies}

\author{Author Names Omitted for Anonymous Review. Paper-ID [317]}

\author{\authorblockN{Chen Xu$^{1}$, Tony Khuong Nguyen$^{1}$, Emma Dixon$^{1}$, Christopher Rodriguez$^{1}$, Patrick Miller$^{1}$, Robert Lee$^{2}$, \\ Paarth Shah$^{1}$, Rares Ambrus$^{1}$, Haruki Nishimura$^{1}$, and Masha Itkina$^{1}$}
\authorblockA{$^{1}$Toyota Research Institute (TRI), $^{2}$Woven by Toyota (WbyT)\\ \texttt{chen.xu@tri.global}}
}


%

\maketitle

\begin{abstract}
Recent years have witnessed impressive robotic manipulation systems driven by advances in imitation learning and generative modeling, such as diffusion- and flow-based approaches. As robot policy performance increases, so does the complexity and time horizon of achievable tasks, inducing unexpected and diverse failure modes that are difficult to predict a priori. To enable trustworthy policy deployment in safety-critical human environments, reliable runtime failure detection becomes important during policy inference. However, most existing failure detection approaches rely on prior knowledge of failure modes and require failure data during training, which imposes a significant challenge in practicality and scalability. In response to these limitations, we present FAIL-Detect, a modular two-stage approach for failure detection in imitation learning-based robotic manipulation. To accurately identify failures from successful training data alone, we frame the problem as sequential out-of-distribution (OOD) detection. We first distill policy inputs and outputs into scalar signals that correlate with policy failures and capture epistemic uncertainty. FAIL-Detect then employs conformal prediction (CP) as a versatile framework for uncertainty quantification with statistical guarantees. Empirically, we thoroughly investigate both learned and post-hoc scalar signal candidates on diverse robotic manipulation tasks. Our experiments show learned signals to be mostly consistently effective, particularly when using our novel flow-based density estimator. Furthermore, our method detects failures more accurately and faster than state-of-the-art (SOTA) failure detection baselines. These results highlight the potential of FAIL-Detect to enhance the safety and reliability of imitation learning-based robotic systems as they progress toward real-world deployment. Videos of FAIL-Detect can be found on our website: \url{https://cxu-tri.github.io/FAIL-Detect-Website/}.
\end{abstract}

\IEEEpeerreviewmaketitle

\section{Introduction}
Robotic manipulation has applications in many important fields, such as manufacturing, logistics, and healthcare \citep{hagele2016industrial}. 
Recently, imitation learning algorithms have shown tremendous success in learning complex manipulation skills from human demonstrations using stochastic generative modeling, such as diffusion-~\citep{chi2023diffusionpolicy,zhao2024aloha} and flow-based methods~\citep{chen2024behavioral,rouxel2024flow}. 
However, despite their outstanding results, policy networks can fail due to poor stochastic sampling from the action distribution.
The models may also encounter out-of-distribution~(OOD) conditions where the input observations deviate from the training data distribution. In such cases, the generated actions may be unreliable or even dangerous. Therefore, it is imperative to detect these failures quickly to ensure the safety and reliability of the robotic system.

\begin{figure*}[!t]
    \centering
    \begin{minipage}[b]{0.15\linewidth}
        \includegraphics[width=\linewidth]{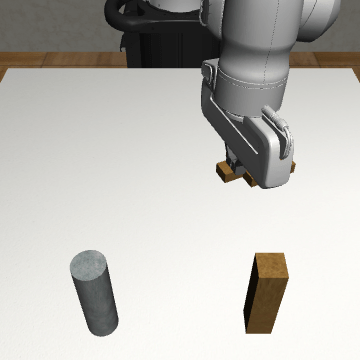}
        \subcaption{Slipped out early.}
    \end{minipage}
    \begin{minipage}[b]{0.15\linewidth}
        \includegraphics[width=\linewidth]{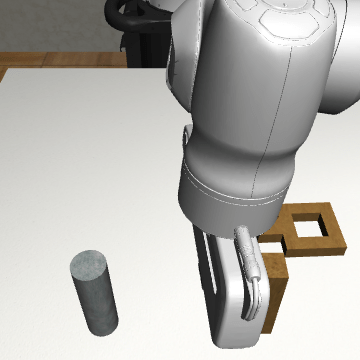}
        \subcaption{Slipped out late.}
    \end{minipage}
    \begin{minipage}[b]{0.15\linewidth}
        \includegraphics[width=\linewidth]{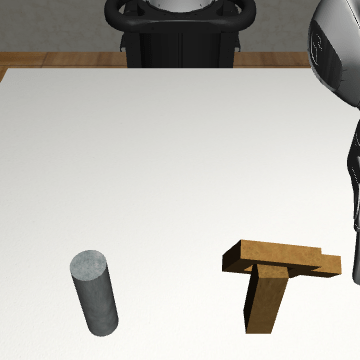}
        \subcaption{Tilted upward.}
    \end{minipage}
    \begin{minipage}[b]{0.15\linewidth}
        \includegraphics[width=\linewidth]{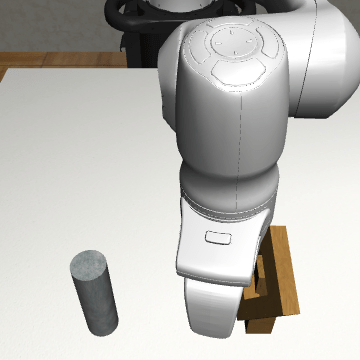}
        \subcaption{Tilted downward.}
    \end{minipage}
    \begin{minipage}[b]{0.15\linewidth}
        \includegraphics[width=\linewidth]{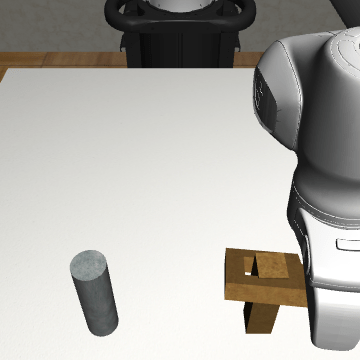}
        \subcaption{Tilted slightly.}
    \end{minipage}
    \begin{minipage}[b]{0.15\linewidth}
        \includegraphics[width=\linewidth]{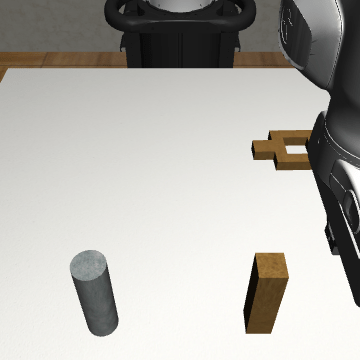}
        \subcaption{Not picked up.}
    \end{minipage}
    \caption{\small  Diverse failure types observed for a single trained policy $g$ on a simple pick-and-place task (put square on peg). These failures occurred at different time steps across multiple rollouts and include the square slipping out of the gripper or being misplaced (e.g., with tilted position) on the peg. FAIL-Detect is able to handle the wide range of failures observed at test time.}
    \label{fig:failure_modes_sim}
\end{figure*}

Detecting failures in robotic manipulation tasks poses several challenges. First, the input data for failure detection, such as environment observations, is often high-dimensional with complicated distributions. This makes it difficult to identify discriminative features that distinguish between successful and failed executions, particularly in the imitation learning setting where a reward function is not defined. Second, there are countless opportunities for failure due to the complex nature of manipulation tasks and the wide range of possible environmental conditions (see \cref{fig:failure_modes_sim}). Consequently, failure detectors must be general and robust to diverse failure scenarios.


\begin{figure*}[!t]
    \centering
    \includegraphics[width=0.95\linewidth]{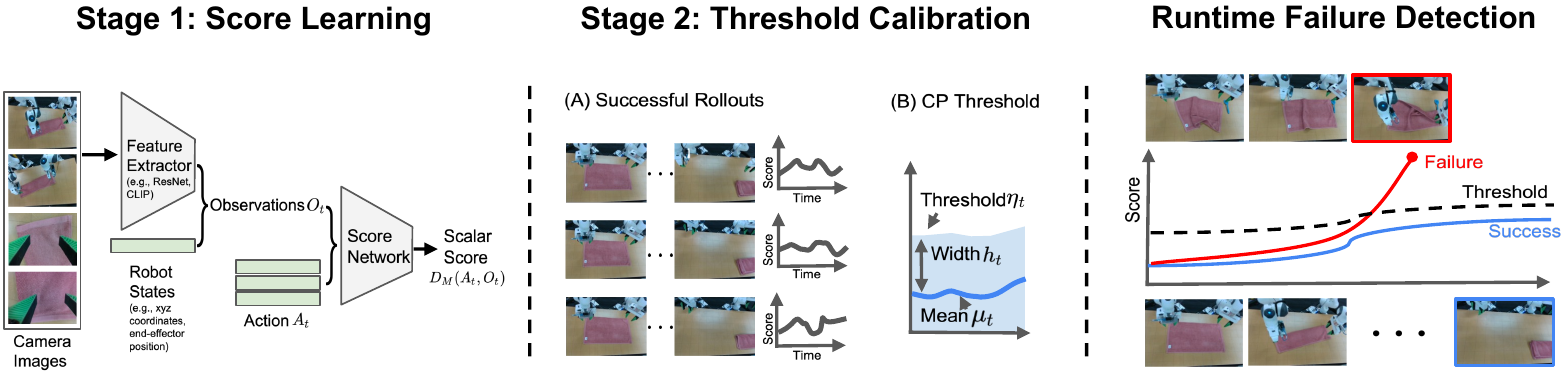}
    \caption{
    \small
    \textbf{FAIL-Detect}: \textbf{F}ailure \textbf{A}nalysis in \textbf{I}mitation \textbf{L}earning – \textbf{Detect}ing failures  without failure data. We propose a two-stage approach to failure detection. 
    \textbf{(Left - Stage I)} Multi-view camera images and robot states are distilled into failure detection scalar scores. Images are first passed through a feature extractor and then, along with robot states, constitute observations $O_t$. Both $O_t$ and generated future robot actions $A_t$ can serve as inputs to a score network $D_M$. This network outputs scalar scores $D_M(A_t, O_t)$ that capture characteristics of successful demonstration data. \textbf{(Middle - Stage II)} Scores from a calibration set of successful rollouts are then used to compute a mean $\mu_t$ and band width $h_t$ to build the time-varying conformal prediction threshold. \textbf{(Right - Runtime Failure Detection)} A successful trajectory (bottom) has scores that consistently remain below the threshold. When a failure occurs (top), such as failure to fold the towel, the score spikes above the threshold, triggering failure detection (red box).
    }
    \label{fig:approach}
    \vspace{-15pt}
\end{figure*}

In imitation learning, training data naturally consists of successful trajectories, making failed trajectories OOD. Prior work often tackles failure detection through binary classification of ID and OOD conditions~\citep{liu2024model}. Thus, many of these methods~\citep{inceoglu2023multimodal,foutter2023selfsupervised,gokmen2023asking,liu2024model} require OOD data for training the failure classifier. This poses significant challenges since collecting and annotating a comprehensive set of failure examples is often time-consuming, expensive, and even infeasible in many real-world scenarios. Moreover, classifiers trained on specific sets of OOD data may not generalize well to unseen failure modes. To address these limitations, we develop a failure detection approach that operates without access to OOD data, overcoming the reliance on failure examples while maintaining robust performance.

We propose \textbf{FAIL-Detect}: \textbf{F}ailure \textbf{A}nalysis in \textbf{I}mitation \textbf{L}earning – \textbf{Detect}ing failures without failure data. FAIL-Detect is a two-stage approach (see \cref{fig:approach}) to failure detection in generative imitation-learning policies. \textbf{In the first stage}, we extract scalar signals from policy inputs and/or outputs (e.g., robot states, visual features, generated future actions) that are discriminative between successes and failures during policy inference. We investigate both learned and post-hoc signal candidates, finding learned signals to be the most accurate for failure detection. A key novelty of our method is the ability to learn failure detection signals without access to failure data. 
 Aside from being performant, our method enables faster inference than prior work~\citep{agia2024unpacking}, which requires sampling multiple robot actions during inference.
\textbf{In the second stage}, we use conformal prediction (CP)~\citep{vovk2005algorithmic,shafer2008tutorial} to construct a time-varying threshold to sequentially determine when a score indicates failure with statistical guarantees on false positive rates. 
By integrating adaptive functional CP~\citep{diquigiovanni2024importance} into our pipeline, we obtain thresholds that adjust to the changing dynamics of manipulation tasks unlike static thresholds used in prior work~\citep{agia2024unpacking}. 

Our contributions are as follows. We present FAIL-Detect, a modular two stage uncertainty-aware runtime failure detection framework for generative imitation learning-based robotic manipulation. First, we construct scalar scores representative of task successes. Second, we use CP to build a time-varying threshold with stochastic guarantees. FAIL-Detect is flexible to incorporate new score and threshold designs. We thoroughly test learned and post-hoc score candidates that rely solely on successful demonstrations. FAIL-Detect with our novel learned score (logpZO) derived from a flow-based density estimator surpasses other methods. We show that FAIL-Detect identifies failures accurately and quickly on diverse robotic manipulation tasks, both in simulation and on robot hardware, outperforming SOTA failure detection baselines.


\section{Related Work}
\textbf{Imitation Learning for Robotic Manipulation.} Imitation learning has emerged as a powerful paradigm for teaching robots complex skills by learning from expert demonstrations. 
Diffusion policy (DP) \citep{chi2023diffusionpolicy} using diffusion models \citep{ho2020denoising} has emerged as highly performant in this space. DP learns to denoise trajectories sampled from a Gaussian distribution, effectively capturing the multi-modal action distributions often present in human demonstrations~\citep{Yu-RSS-23,chi2023diffusionpolicy}. Diffusion models have been used to learn observation-conditioned policies~\citep{chi2023diffusionpolicy,sridhar2024nomad}, integrate semantic information via language conditioning~\citep{Yu-RSS-23,chen2023playfusion,reuss2023multimodal}, and improve robustness and generalization~\citep{kim2024robust,wang2024equivariant,wang2024gendp}. Concurrently, vision-language-action models like Octo \citep{octo_2023} and OpenVLA \citep{kim24openvla} have shown promise in generalist robot manipulation by leveraging large-scale pretraining on diverse datasets. More recently, flow matching (FM) generative models have been proposed as an alternative to diffusion in imitation learning~\citep{black2024pi0,hu2024adaflow,braun2024riemannian,zhang2024flowpolicy}, offering faster inference and greater flexibility (i.e., extending beyond Gaussian priors~\citep{chen2024behavioral}), while achieving competitive or superior success rates. We test FAIL-Detect on DP and FM imitation learning architectures.

\textbf{OOD Detection.} The task of detecting robot failures can be viewed as anomaly detection, which falls under the broader framework of OOD detection~\citep{yang2024generalized}. 
Ensemble methods \citep{lakshminarayanan2017simple}, which combine predictions from multiple models to improve robustness and estimate uncertainty, have long been regarded as the de facto approach for addressing this problem. However, they are computationally expensive as they require training and running inference on multiple models.
Another popular approach frames OOD detection as a classification problem~\citep{liu2024multitask}.
This formulation learns the decision boundary between ID and OOD data by training a binary classifier~\citep{vyas2018out,djurisic2023extremely}, but requires OOD data during training.
In contrast, density-based approaches~\citep{liu2020energy,du2022towards,xu2023normalizing}, one-class discriminators based on random networks~\citep{Ciosek2020Conservative,He2024}, and control-theoretic methods~\citep{castaneda2023idbf} aim to model information from ID data without relying on OOD data during training. Density-based methods attempt to capture the distribution of ID data, yet they can be challenging to optimize. One-class discriminators have shown superior performance over deep ensembles in practice but can be sensitive to the design of the discriminator model. Control-theoretic approaches use contrastive energy-based models; however, they often require a representation of the system's dynamics.
Furthermore, evidential deep learning methods~\citep{charpentier2022natural,Bramlage2023,itkina2023interpretable} learn parameters for second-order distributions (e.g., Dirichlet) to approximate epistemic uncertainty (due to limited model knowledge or OOD inputs) from aleatoric uncertainty (due to inherent randomness in the data).
Lastly, distance-based approaches \citep{basart2022scaling,tao2023nonparametric,kaur2023codit} identify OOD samples by computing their distance to ID samples in the input or latent space, avoiding the need for training but exhibiting limited performance compared to other approaches. We consider many of the listed model variants as score candidates in FAIL-Detect.

\textbf{Failure Detection in Robotics.} Detecting failures in robotic systems is important for ensuring safety and reliability, as failures can lead to undesirable behaviors in human environments~\citep{NATARAJAN2023100298,9336665,liu2024multitask}. Various approaches have been proposed, such as building fast anomaly classifiers based on LLM embeddings~\citep{Sinha-RSS-24} and using the reconstruction error from variational autoencoders (VAE) to detect anomalies in behavior cloning (BC) policies for mobile manipulation~\citep{wong2022error}. Separately, \citet{ren2023robots} construct uncertainty sets from conformal prediction for actions generated by an LLM-based planner, prompting human intervention when the set is ambiguous. These works do not consider failure detection in the setting of generative imitation learning policies. 
On the other hand, \citet{gokmen2023asking} learn a state value function that is trained jointly with a BC policy and can be used to predict failures. \citet{liu2024model} propose an LSTM-based failure classifier for a BC-RNN policy using latent embeddings from a conditional VAE. Given a Transformer-based policy and a world model to predict future latent embeddings, \citet{liu2024multitask} 
train a failure detection classifier on the embeddings. To handle previously unseen states, they also propose a SOTA OOD detection method, which we adapt as a baseline to our approach (see PCA-kmeans in \cref{sec:experiments}). However, unlike FAIL-Detect, these methods require collecting failed trajectories a priori to detect failures. Meanwhile, \citet{wang2024grounding} uses self-reset to collect additional failure data and train a classifier to identify failure modes. In their on-robot experiments, approximately 2000 trajectories (roughly 2 hours) had to be collected using self-reset, making scalability challenging. For diffusion-based policies, \citet{sun2023conformal} reduce model uncertainty by producing prediction intervals for rewards of predicted trajectories. \citet{He2024} propose using random network distillation (RND) to detect OOD trajectories and select reliable ones. These works do not directly consider runtime failure detection. Our two-stage solution for this problem combines the advantages of both approaches. The closest SOTA method to FAIL-Detect by \citet{agia2024unpacking} introduces a statistical temporal action consistency (STAC) measure in conjunction with vision-language models (VLMs) to detect failures within rollouts at runtime. STAC does not require failure data, consists of a score computed post-hoc from a batch of predicted actions and a constant-time CP threshold to flag failures, and is evaluated in the context of DP. We demonstrate improved empirical performance over STAC by integrating \textit{learned} failure detection scores with a \textit{time-varying} CP band.
\begin{table*}[!t]
\centering
\caption{Overview of score methods evaluated in this work. The input was selected either based on the structure and requirements of each method or, when multiple input combinations were possible, based on empirical performance. All methods
except STAC (which proposes a different calibration method; see \cref{sec:experiments}) use time-varying CP bands described in \cref{sec:threshold}.}
\label{tab:score_overview}
\resizebox{\textwidth}{!}{
\begin{tabular}{llllll}
\toprule
Method & Type & Input & Category & Novelty & Original application \\
\midrule
logpZO & Learned & $O_t$ & Density estimation & Novel & N/A \\
lopO & Learned & $O_t$ & Density estimation & Adapted~\citep{xu2023normalizing} & Likelihood estimation on tabular data \\
NatPN & Learned & $O_t$ & Second-order & Adapted~\citep{charpentier2022natural} & OOD detection for classification and regression \\
DER & Learned & $(O_t, A_t)$ & Second-order & Adapted~\citep{Bramlage2023} & OOD detection for human pose estimation\\
RND & Learned & $(O_t, A_t)$ & One-class discriminator & Adapted~\citep{He2024} &  Reinforcement learning \citep{He2024}; OOD detetcion \citep{Ciosek2020Conservative} \\
CFM & Learned & $O_t$ & One-class discriminator & Adapted~\citep{yang2024consistency} & Efficient sampling of flow models \\
SPARC & Post-hoc & $A_t$ & Smoothness measure & Adapted~\citep{balasubramanian2015analysis}  & Smoothness analysis for time series data \\
STAC & Post-hoc & $A_t$ & Statistical divergence & Baseline~\citep{agia2024unpacking} & Failure detection for generative imitation learning policies \\
PCA-kmeans & Post-hoc & $O_t$ & Clustering & Baseline~\citep{liu2024multitask} & OOD detection during robot execution \\
\bottomrule
\end{tabular}}
\end{table*}

\section{Problem Formulation}
Our focus in this work is to detect when a generative imitation learning policy fails to complete its task during execution. We define the following notation. Let $g(A_t \mid O_t)$ denote the generator, where $O_t$ represents the environment observation (e.g., image features and robot states) at time~$t$, and $g$ is a stochastic predictor of a sequence of actions $A_t=(A_{t \mid t}, A_{t+1 \mid t}, \ldots, A_{t+H-1 \mid t})$ for the next $H$ time steps. 
The first $H' < H$ actions $A_{t:t+H' \mid t}$ are executed, after which the robot re-plans by generating a new sequence of $H$ actions at time $t+H'$. Recent works have trained effective generators~$g$ via DP~\citep{chi2023diffusionpolicy} and FM~\citep{chen2024behavioral}. Given an initial condition~$O_0$ and the generator~$g$ to output the next actions, we obtain a trajectory $\tau_t=(O_0, A_0, O_{H'}, A_{H'}, \ldots, O_t, A_t)$ up to $t=kH'$ ($k\geq 1$) execution time steps. Failure detection can thus be framed as designing a decision function $D(\cdot;\theta):\tau_t \rightarrow \{0,1\}$ with parameters $\theta$, which takes in the current trajectory and makes a decision. If the decision $D(\tau_t;\theta)=1$, the rollout is flagged as a failure at time step $t$. For instance, in a pick-and-place task, a failure may be detected after the robot fails to pick up the object or misses the target position.

\section{Failure Detection Framework}
\label{sec:methods}

Given action-observation data $(A_t,O_t)$, we propose a two-stage framework to design the decision function $D(\cdot;\theta)$:
\begin{enumerate}
\item Train a scalar score model $D_M(\cdot;\theta): (A_t, O_t) \rightarrow \mathbb{R}$ (for score ``method'' $M$) on action and/or observation pairs from successful trajectories only. 
\item Calibrate time-varying thresholds $\eta_t$ based on a CP band. 
\end{enumerate}
The final decision $D(\tau_t;\theta)=\mathbbm{1}(D_M(A_t,O_t;\theta)>\eta_t)$ raises a failure flag if the scalar score $D_M(A_t,O_t;\theta)$ exceeds the threshold $\eta_t$ at time step $t$. This two-stage framework is flexible to incorporate new scores in Stage 1 or new thresholds in Stage 2.
See \cref{fig:approach} for an overview of the framework.

\subsection{Design of Scalar Scores}\label{sec:scalar_extractor}

To construct scores indicative of failures, we propose a novel score candidate and several adaptations of existing approaches originally developed for other applications. See \cref{tab:score_overview} for an overview of the scoring methods we consider.

When designing a scalar score that is indicative of policy failure, we consider the following desiderata:
\textbf{(1)} \textbf{One-class}: The method should not require failure data during training as it 
may be too diverse to enumerate (see \cref{fig:failure_modes_sim}).
\textbf{(2)} \textbf{Light-weight}: The method should allow for fast inference to enable real-time robot manipulation.
\textbf{(3)} \textbf{Discriminative}: The method should yield gaps in scores for successful and failed rollouts.
To avoid overfitting on historical data, the score network $D_M$ only takes the latest $T_O$ steps ($T_O=2$ following \citep{chi2023diffusionpolicy}) of past observations $O_t$ alongside future action $A_t$ as inputs, rather than the growing trajectory history. To meet our desiderata, we select and build on the following approach categories.

\textbf{(a)} \textbf{Learned data density:} we fit a normalizing flow-based density estimator to the observations, where data far from the distribution of successful trajectory observations may indicate failure. 
The approach we term lopO~\citep{xu2023normalizing} fits a continuous normalizing flow (CNF) $f_{\theta}$ to the set of observations $\{O_t\}_{t\geq 0}$. A low $\log p(O_{t'})$ for a new observation $O_{t'}$ implies it is unlikely, indicating possible failure. Note the computation of $\log p(O_{t'})$ requires integration of the divergence of $f_{\theta}$ over the ODE trajectory, which is difficult to estimate in high dimensions.
Additionally, we introduce our novel logpZO approach, which leverages the same CNF $f_{\theta}$ to evaluate the likelihood of a noise estimate $Z_{O_{t}}$ (conditioned on an observation $O_{t}$). Using the forward ODE process, we compute $Z_{O_{t}}$ by integrating $f_{\theta}$ over the unit interval [0,1], starting from $O_t$ as the ODE initial condition. When $O_{t}$ is ID, $Z_{O_{t}}$ is approximately Gaussian, leading to $p(Z_{O_{t}}) = C\exp(-0.5|Z_{O_{t}}|^2)$. Thus, a high value of $|Z_{O_{t}}|_2^2$ corresponds to a low likelihood $p(Z_{O_{t}})$ in the noise space. Further details on logpZO are described in \cref{appendix:logpZO}. The key distinction between lopO and logpZO lies in their domains: the former assesses likelihood in the original observation space, while the latter does so in the latent noise space. We expect the latter to be better because its computation does not require the divergence of $f_{\theta}$ integrated over $[0,1]$, a hard-to-estimate quantity in high dimensions.

\textbf{(b)} \textbf{Second-order:} these methods learn parameters for second-order distributions that can separate aleatoric and epistemic uncertainty~\cite{sensoy2018evidential}. 
NatPN~\citep{charpentier2022natural} imposes a Dirichlet prior on class probabilities and optimizes model parameters by minimizing a Bayesian loss. To use NatPN, we discretize the observations $O_t$ using $K$-means and apply NatPN to the discretized version.
We also consider multivariate deep evidential regression DER~\citep{Bramlage2023}, which assumes $A_t | O_t$ follows a multivariate Gaussian distribution with a Wishart prior and learns its parameters.

\begin{figure*}[!t]
    \centering
    \begin{minipage}{0.23\linewidth}
        \includegraphics[width=\linewidth]{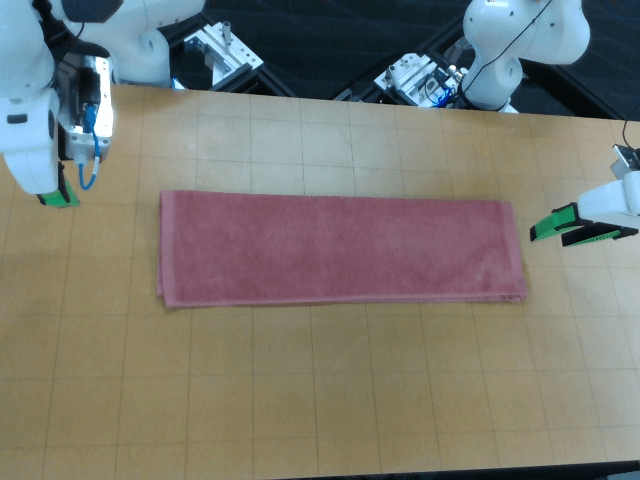}
        \vspace{-15pt}
        \subcaption{Before disturbance}
    \end{minipage}
    \begin{minipage}{0.23\linewidth}
        \includegraphics[width=\linewidth]{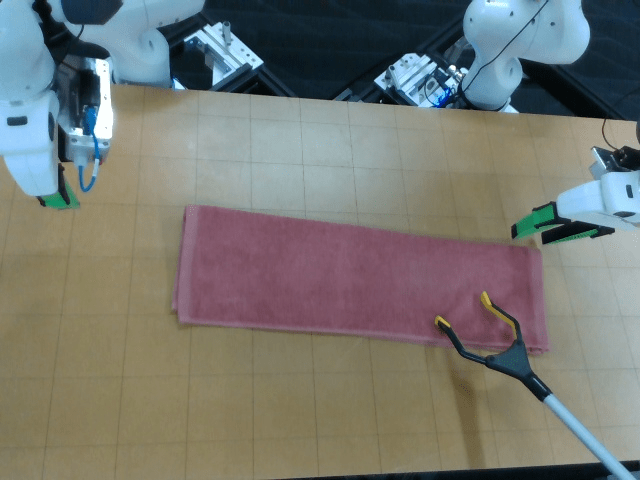}
        \vspace{-15pt}
        \subcaption{After disturbance}
    \end{minipage}
    \begin{minipage}{0.23\linewidth}
        \includegraphics[width=\linewidth]{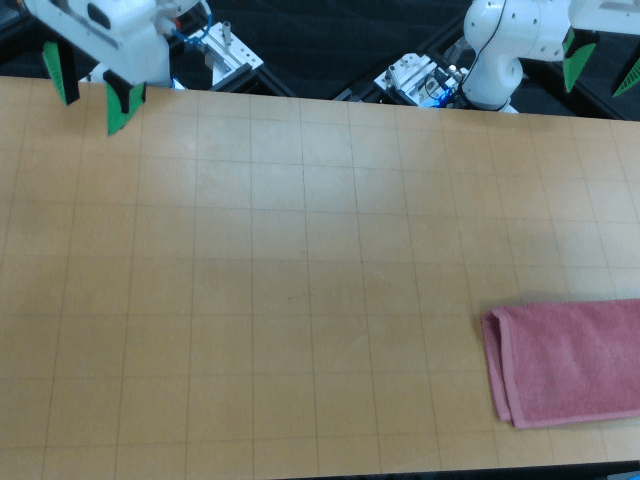}
        \vspace{-15pt}
        \subcaption{Successful rollout}
    \end{minipage}
    \begin{minipage}{0.23\linewidth}
        \includegraphics[width=\linewidth]{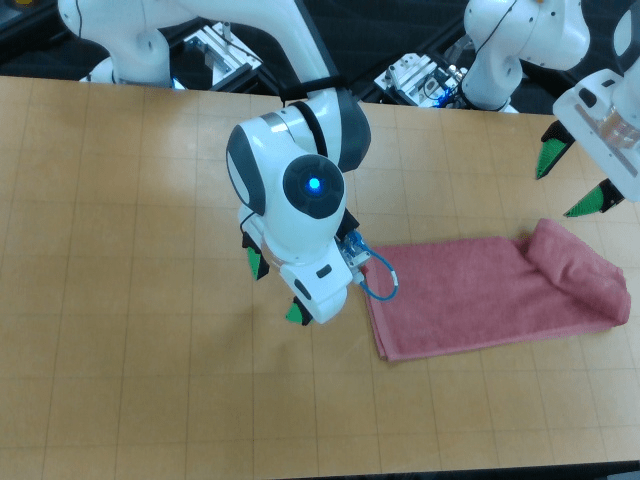}
        \vspace{-15pt}
        \subcaption{Failed rollout}
    \end{minipage}
    \begin{minipage}{0.23\linewidth}
        \includegraphics[width=\linewidth]{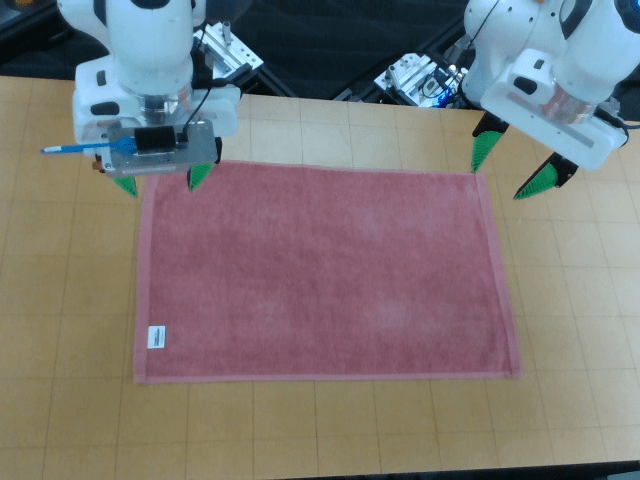}
        \vspace{-15pt}
        \subcaption{ID initial condition}
    \end{minipage}
    \begin{minipage}{0.23\linewidth}
        \includegraphics[width=\linewidth]{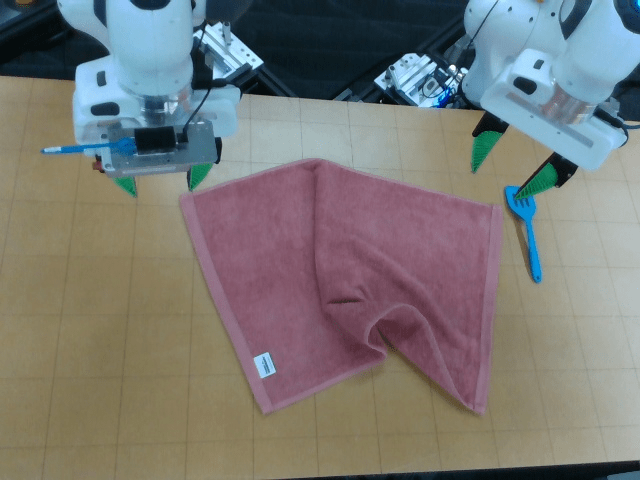}
        \vspace{-15pt}
        \subcaption{OOD initial condition}
    \end{minipage}
    \begin{minipage}{0.23\linewidth}
        \includegraphics[width=\linewidth]{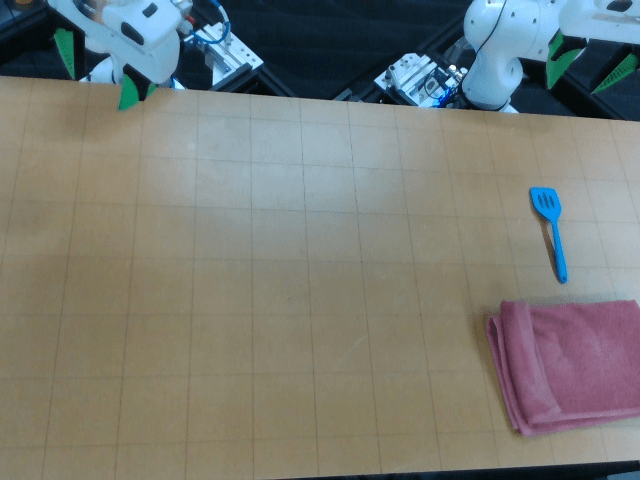}
        \vspace{-15pt}
        \subcaption{Successful rollout}
    \end{minipage}
    \begin{minipage}{0.23\linewidth}
        \includegraphics[width=\linewidth]{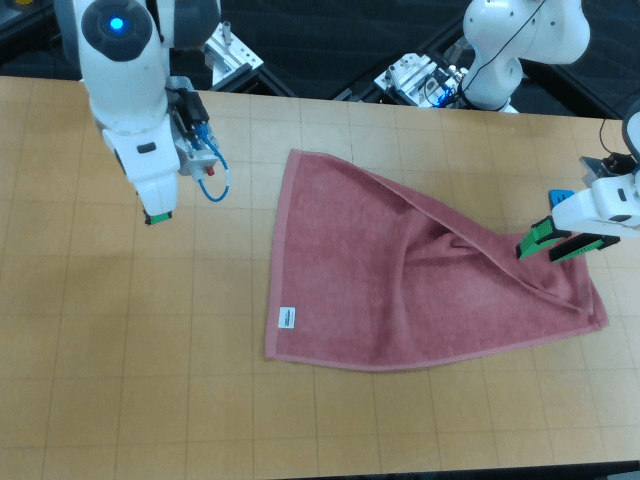}
        \vspace{-15pt}
        \subcaption{Failed rollout}
    \end{minipage}

    \begin{minipage}{0.23\linewidth}
        \includegraphics[width=\linewidth]{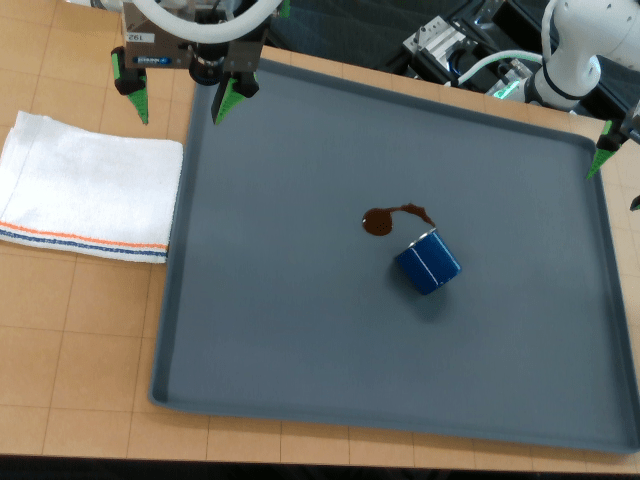}
        \vspace{-15pt}
        \subcaption{ID initial condition}
    \end{minipage}
    \begin{minipage}{0.23\linewidth}
        \includegraphics[width=\linewidth]{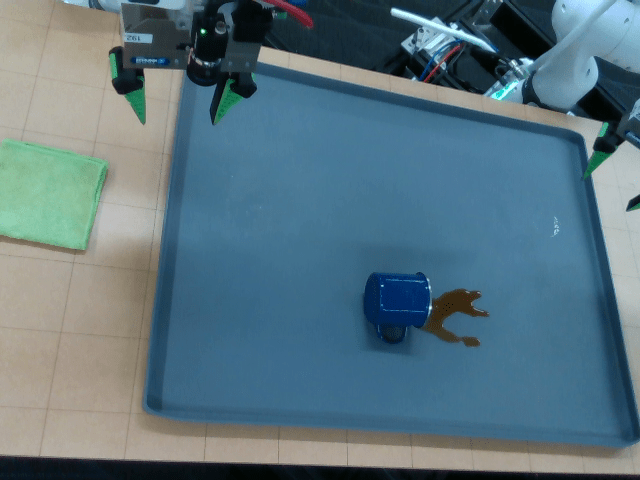}
        \vspace{-15pt}
        \subcaption{OOD initial condition}
    \end{minipage}
    \begin{minipage}{0.23\linewidth}
        \includegraphics[width=\linewidth]{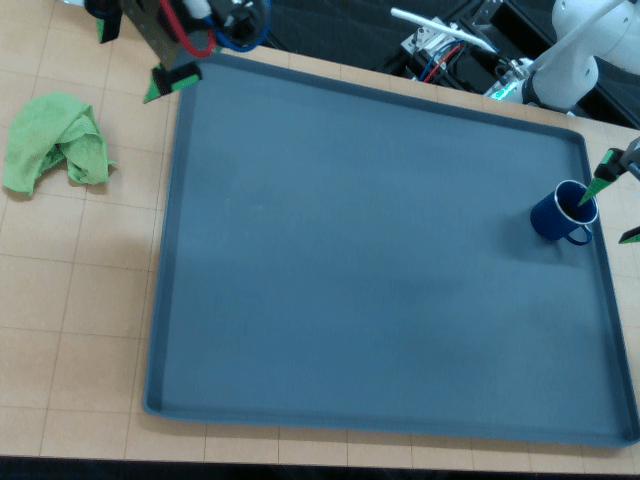}
        \vspace{-15pt}
        \subcaption{Successful rollout}
    \end{minipage}
    \begin{minipage}{0.23\linewidth}
        \includegraphics[width=\linewidth]{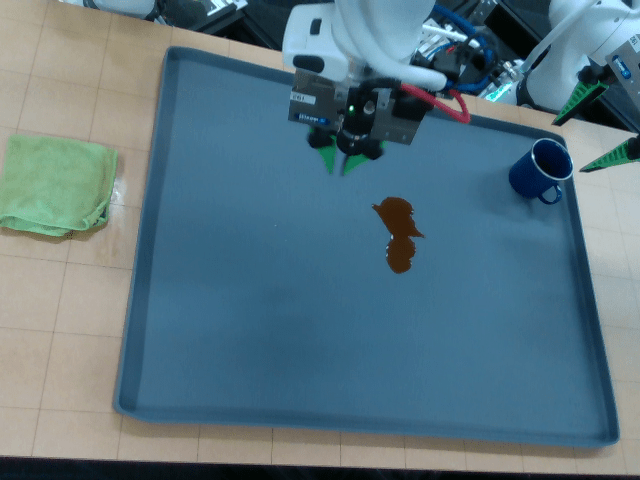}
        \vspace{-15pt}
        \subcaption{Failed rollout}
    \end{minipage}
    \caption{\small  Robot hardware experiment scenarios.
    \textbf{(Top row)} \textbf{FoldRedTowel} with Disturbance: In (b), the human pulls the towel from the position in (a) towards the bottom during a policy rollout. We note that such recovery behavior is sometimes present in the training data, so the task may succeed as in (c). A failure case is shown in (d).
    \textbf{(Middle row)} \textbf{FoldRedTowel} OOD: Compared to ID (e), we start with a crumpled towel with a blue spatula distractor to the right of the towel as in (f). Neither condition is present in the training data, thus although the task could succeed as in (g), the success rate is low and the robot typically fails like in (h). 
    \textbf{(Bottom row)} \textbf{CleanUpSpill} OOD: Compared to ID (i), we start with a green towel as in (j). The training data only contains white and gray towels and, therefore, although the task could succeed as in (k), the robot typically fails like in (l) with a low success rate.}
    \label{fig:real_OOD}
    \vspace{-10pt}
\end{figure*}

\textbf{(c)} \textbf{One-class discriminator:} we consider methods that learn a continuous metric, but do not directly model the distribution of input data.
The one-class discriminator RND~\citep{He2024} initializes random target $f_T(\cdot)$ and random predictor $f(\cdot;\theta)$ networks. The target is frozen, while the predictor is trained to minimize $\mathbb{E}_{(A_t,O_t)\sim \text{ID trajectory}} [D_M(A_t,O_t;\theta)]$ for $D_M(A_t,O_t;\theta)=\vert\vert f_T(A_t,O_t)-f(A_t,O_t;\theta)\vert \vert ^2_2$ on successful demonstration data. 
Intuitively, RND learns a mapping from the data $(A_t,O_t)$ to a preset random function. If the learned mapping starts to deviate from the expected random output, the input data is likely OOD.
In this category, we also consider consistency flow matching~(CFM)~\citep{yang2024consistency}, which measures trajectory curvature with empirical variance of the observation-to-noise forward flow. The intuition is that on ID data, the forward flow is trained to be straight and consistent. Thus, high trajectory curvature indicates the input data is OOD.

\textbf{(d)} \textbf{Post-hoc metrics:} we investigate methods that compute a scalar score analytically without learning. We use SPARC~\citep{balasubramanian2015analysis} to measure the smoothness of predicted actions. We expect SPARC to be useful for robot jitter failures, which are empirically frequent in OOD scenarios. The recent SOTA in success-based failure detection, STAC~\citep{agia2024unpacking}, falls in the post-hoc method category. However, since it comes with its own statistical evaluation procedure, we describe it as one of our main baselines in \cref{sec:experiments}. Additionally, we term the OOD detection method by \citet{liu2024multitask} as PCA-kmeans, which also falls in this category. We retrofit it within our two-stage framework as another baseline in \cref{sec:experiments}.

\begin{figure*}[!t]
    \centering
    \begin{minipage}[b]{0.495\linewidth}
        \includegraphics[width=\linewidth]{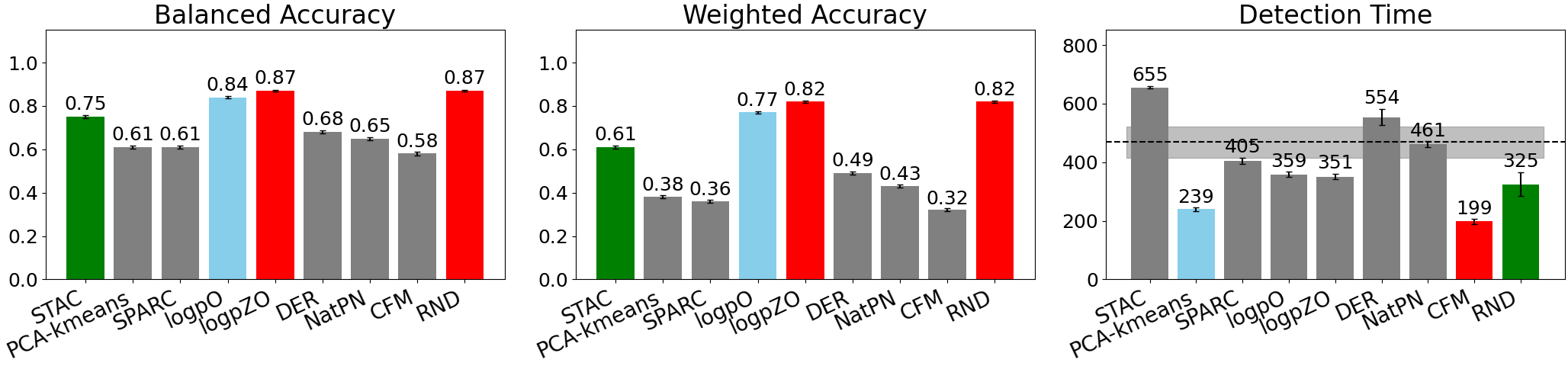}
        \vspace{-20pt}
        \subcaption{Transport ID}
        \vspace{5pt}
    \end{minipage}
    \begin{minipage}[b]{0.495\linewidth}
        \includegraphics[width=\linewidth]{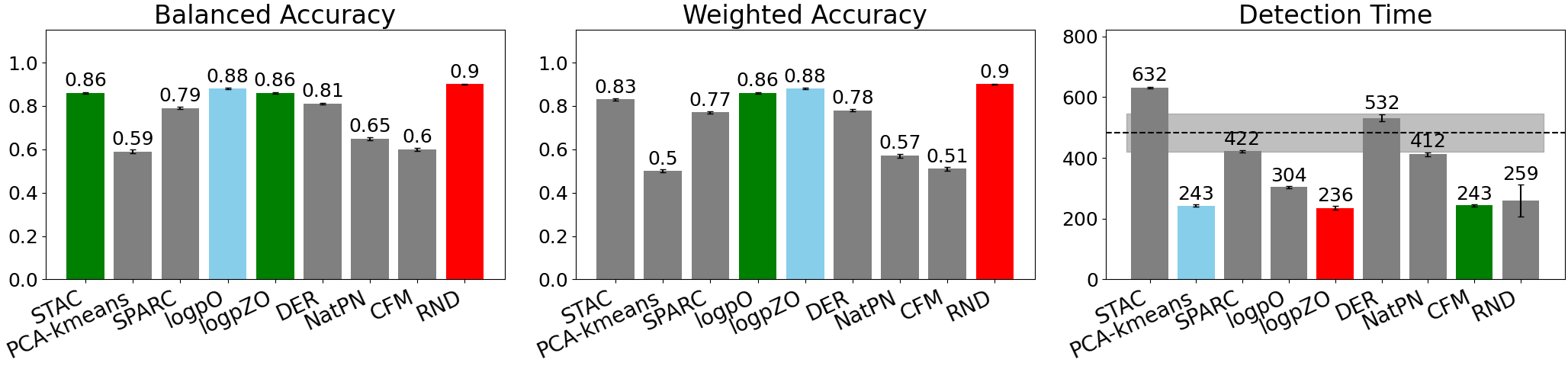}
        \vspace{-20pt}
        \subcaption{Transport OOD}
        \vspace{5pt}
    \end{minipage}
    \begin{minipage}[b]{0.495\linewidth}
        \includegraphics[width=\linewidth]{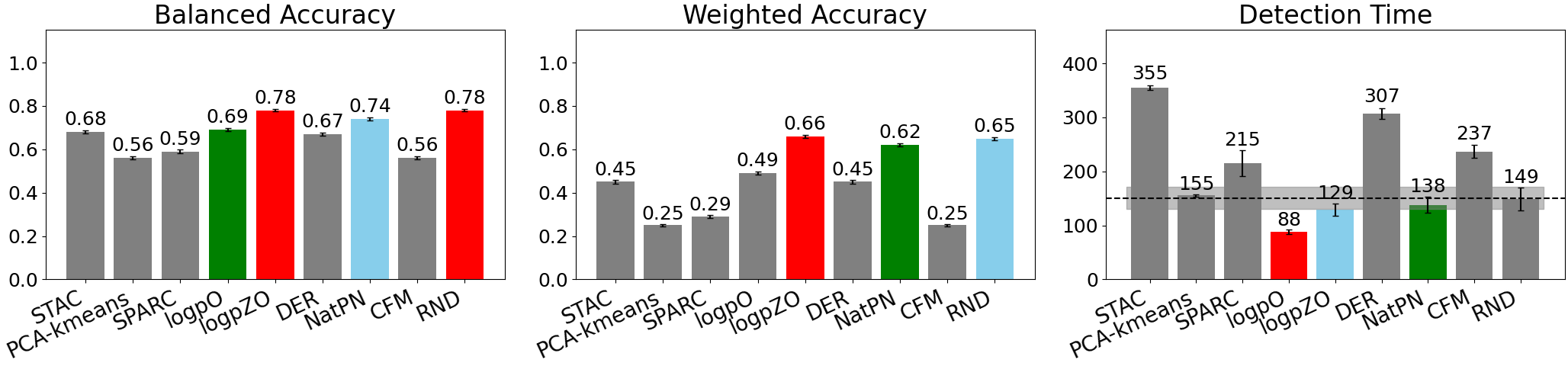}
        \vspace{-20pt}
        \subcaption{Square ID}
        \vspace{5pt}
    \end{minipage}
    \begin{minipage}[b]{0.495\linewidth}
        \includegraphics[width=\linewidth]{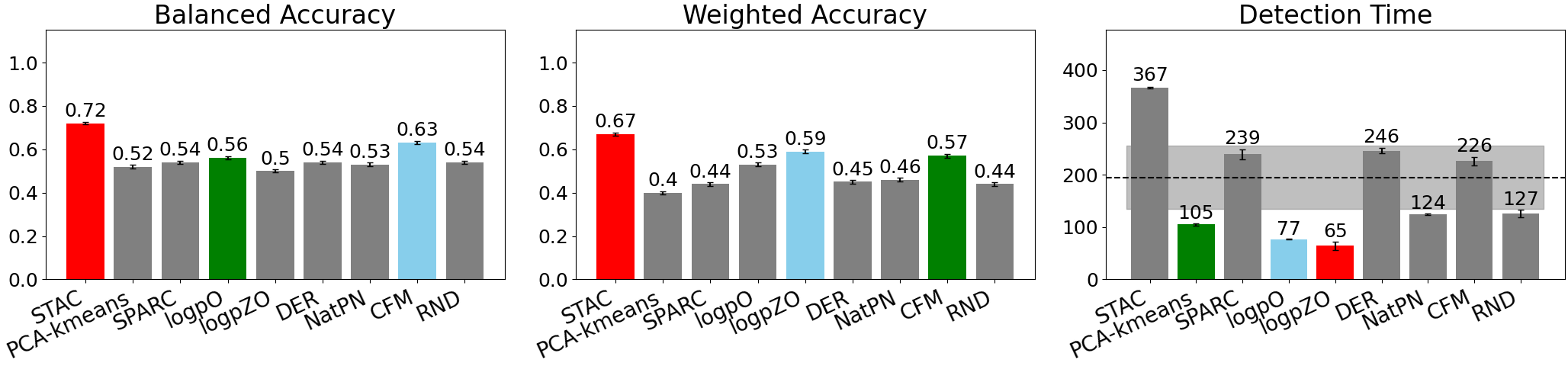}
        \vspace{-20pt}
        \subcaption{Square OOD}
        \vspace{5pt}
    \end{minipage}
    \begin{minipage}[b]{0.495\linewidth}
        \includegraphics[width=\linewidth]{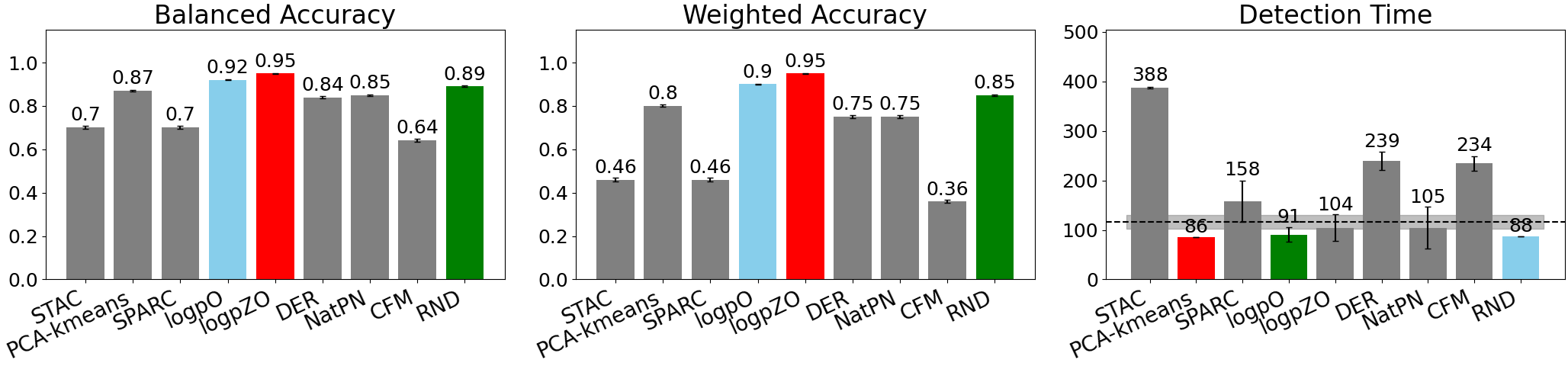}
        \vspace{-20pt}
        \subcaption{Can ID}\label{fig:can_ID}
        \vspace{5pt}
    \end{minipage}
    \begin{minipage}[b]{0.495\linewidth}
        \includegraphics[width=\linewidth]{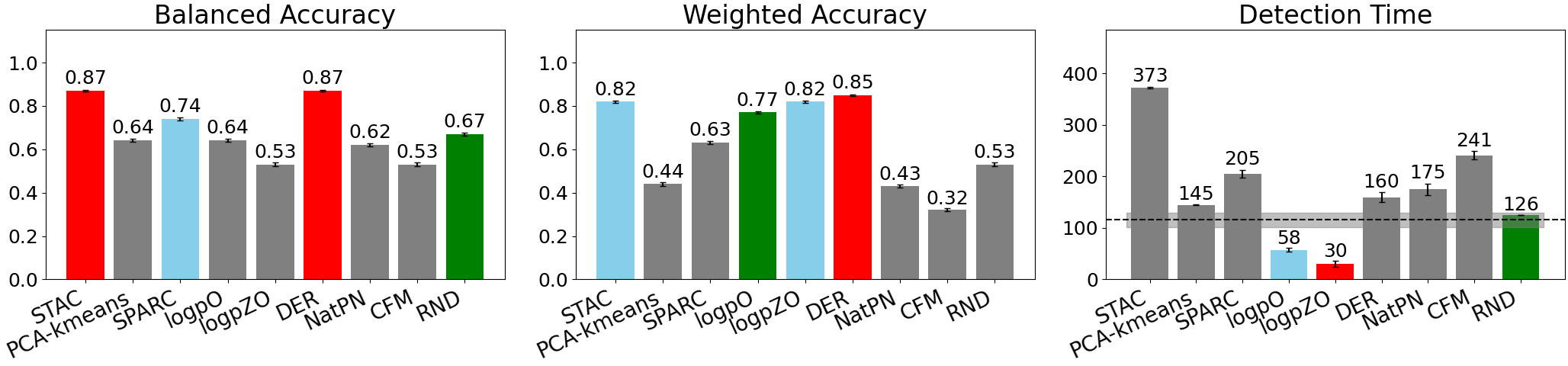}
        \vspace{-20pt}
        \subcaption{Can OOD}
        \vspace{5pt}
    \end{minipage}
    \begin{minipage}[b]{0.495\linewidth}
        \includegraphics[width=\linewidth]{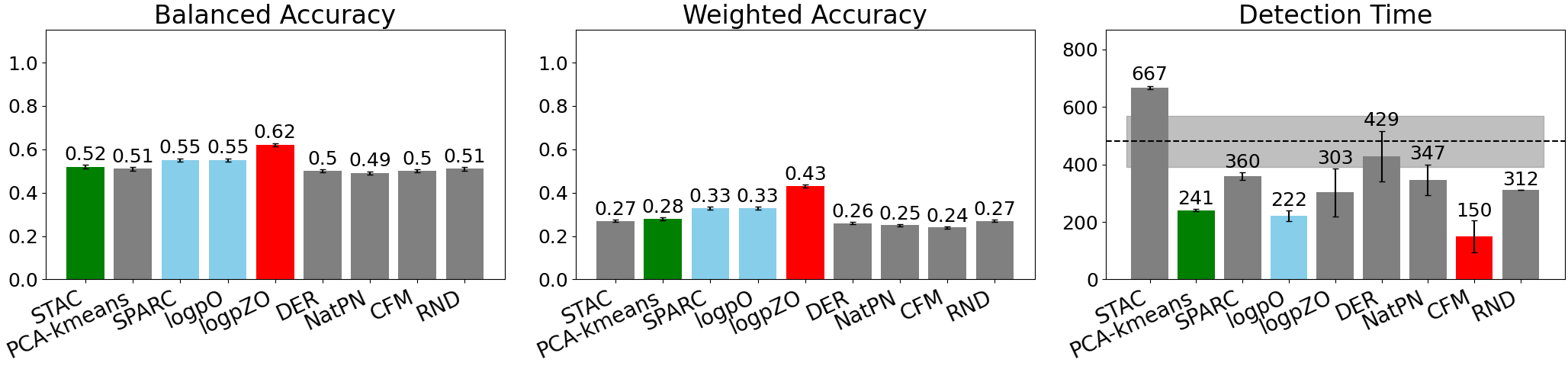}
        \vspace{-20pt}
        \subcaption{Toolhang ID}
    \end{minipage}
    \begin{minipage}[b]{0.495\linewidth}
        \includegraphics[width=\linewidth]{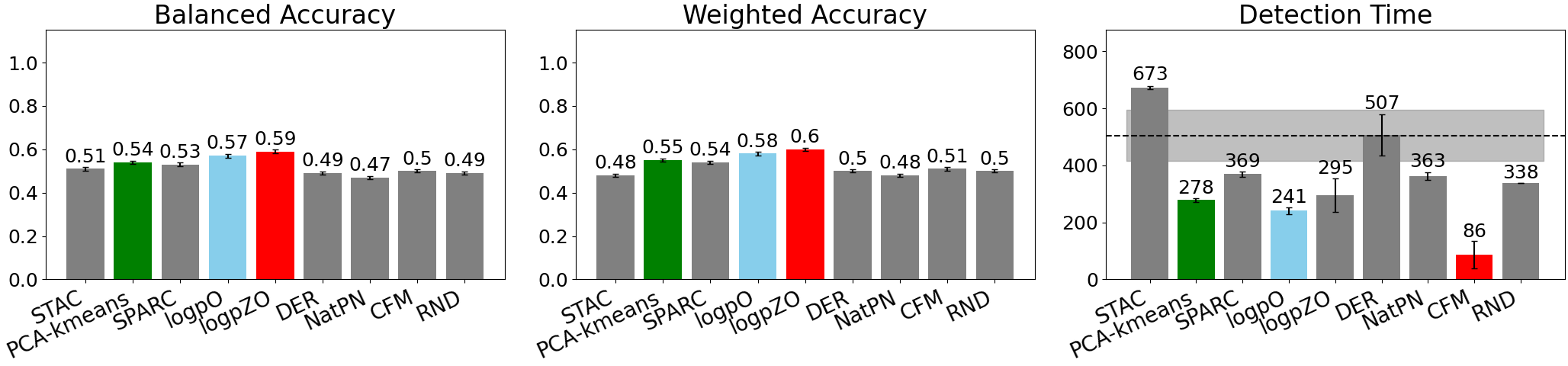}
        \vspace{-20pt}
        \subcaption{Toolhang OOD}
    \end{minipage}
    \caption{\small Quantitative failure detection results for simulation tasks on FM policy ({\color{red} best}, {\color{cyan} second}, {\textcolor{ForestGreen}{third}}); results with TPR and TNR are in \cref{fig:sim_metrics} and results on DP are in \cref{fig:sim_metric_DP}. For balanced accuracy and weighted accuracy, higher is better and for detection time, lower is better. The CP band for each task is calibrated with successful rollouts under ID initial conditions only (i.e., the same band is used for ID and OOD test cases).
    We group together post-hoc (STAC, PCA-kmeans, SPARC), density-based (logpO, logpZO), second-order (DER, NatPN), and one-class (CFM, RND) methods and show barplots with standard errors.
    The dashed line in the Detection Time plots represents the average successful trajectory time in that setting with standard error. 
    Overall, learned methods outperform post-hoc ones in failure detection. 
    In terms of \textbf{combined accuracy} (balanced accuracy and weighted accuracy), logpZO and RND are the best two methods, reaching top-\rev{1 } performance in \rev{10}/16 and \rev{5}/16 cases, respectively. Moreover, logpZO reaches top-\rev{3 } performance in \rev{14}/16 cases, while RND does so in \rev{9}/16 cases. In comparison, the baselines STAC and PCA-kmeans reach top-\rev{1 } performance in \rev{3}/16 and \rev{0}/16 cases, respectively. Note that STAC reaches top-\rev{3 } performance in \rev{8}/16 cases, while PCA-kmeans does so in \rev{3}/16 cases.
    The learned methods also achieve the fastest \textbf{detection time}, with one of the learned methods always getting the best overall detection time in all \rev{but one }case. In terms of best top-\rev{1 } performance, logpZO \rev{is the fastest method } in \rev{3}/8 cases, \rev{RND in 0/8 cases, and } the PCA-kmeans baseline \rev{does so }in \rev{1}/8 cases. In contrast, STAC is the slowest in nearly all cases, detecting failures only after the average success trajectory time, rendering the detection not practical.
    }
    \label{fig:sim_metrics_abbrev}
\end{figure*}

\begin{figure*}[!t]
    \centering
    \begin{minipage}[b]{0.495\linewidth}
        \includegraphics[width=\linewidth]{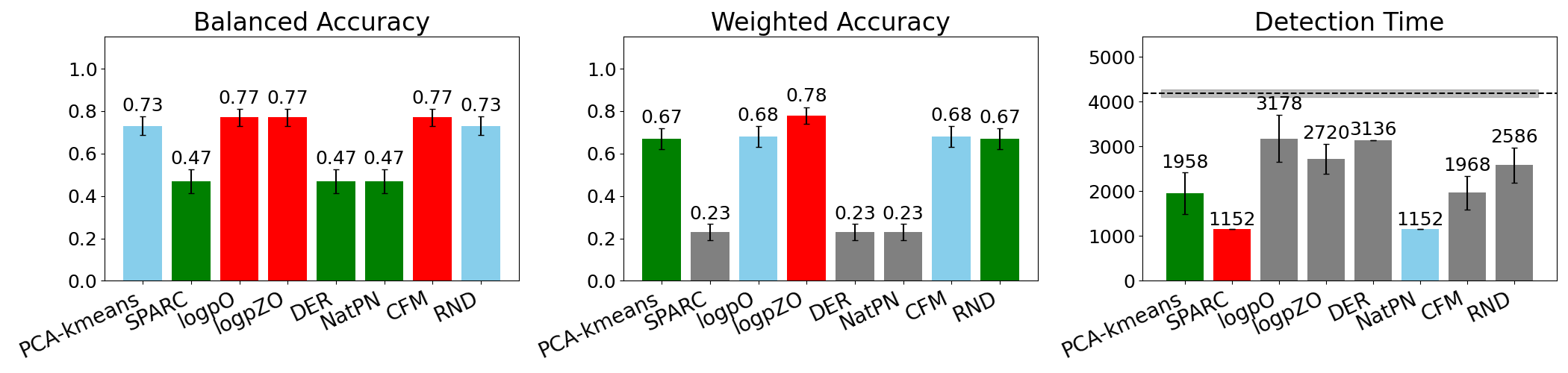}
        \vspace{-20pt}
        \subcaption{\makecell[l]{\textbf{FoldRedTowel} with FM: (Setting-dependent band) ID + Disturb}}
        \vspace{5pt}
    \end{minipage}
    \begin{minipage}[b]{0.495\linewidth}
        \includegraphics[width=\linewidth]{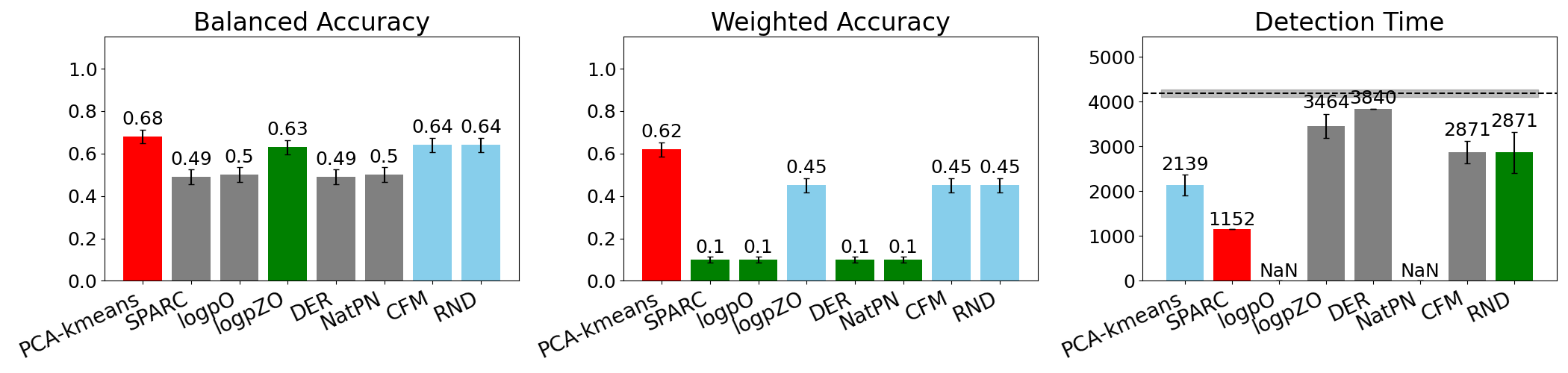}
        \vspace{-20pt}
        \subcaption{\makecell[l]{\textbf{FoldRedTowel} with FM: (ID-only band) ID + Disturb}}
        \vspace{5pt}
    \end{minipage}
    \begin{minipage}[b]{0.495\linewidth}
        \includegraphics[width=\linewidth]{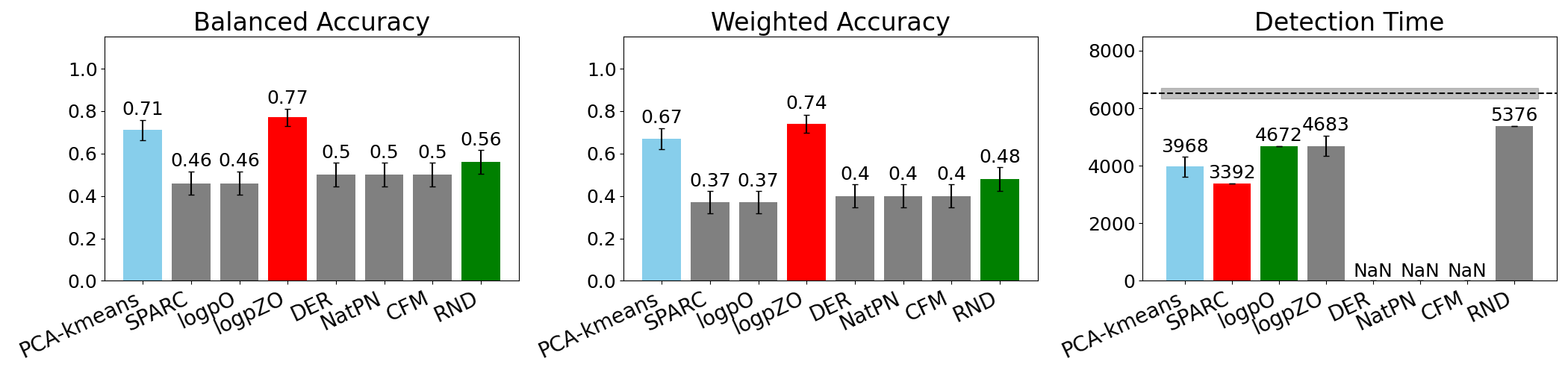}
        \vspace{-20pt}
        \subcaption{\makecell[l]{\textbf{FoldRedTowel} with FM: (Setting-dependent band) OOD}}
        \vspace{5pt}
    \end{minipage}
    \begin{minipage}[b]{0.495\linewidth}
        \includegraphics[width=\linewidth]{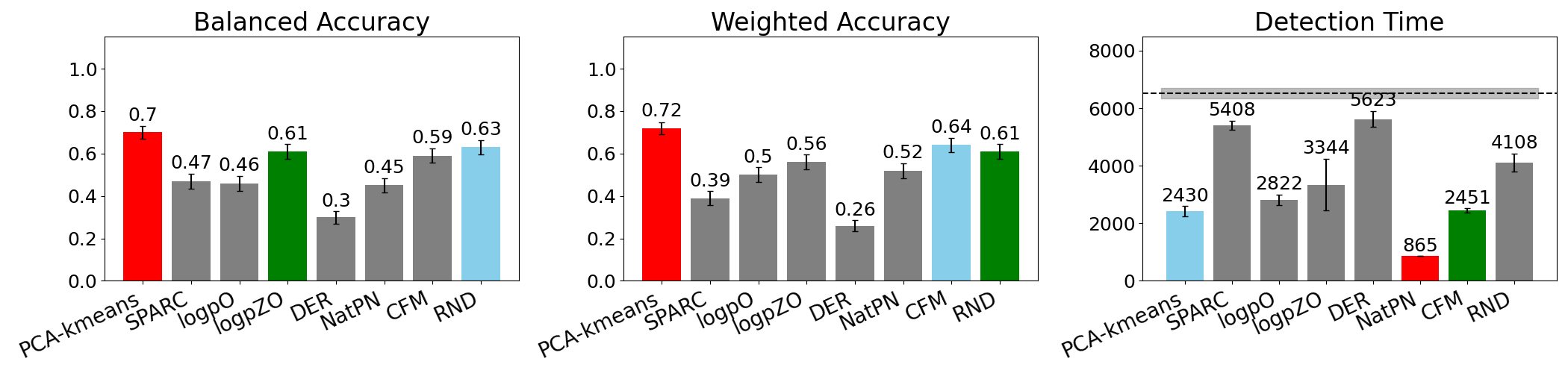}
        \vspace{-20pt}
        \subcaption{\makecell[l]{\textbf{FoldRedTowel} with FM: (ID-only band) OOD}}
        \vspace{5pt}
    \end{minipage}
    \begin{minipage}[b]{0.495\linewidth}
        \includegraphics[width=\linewidth]{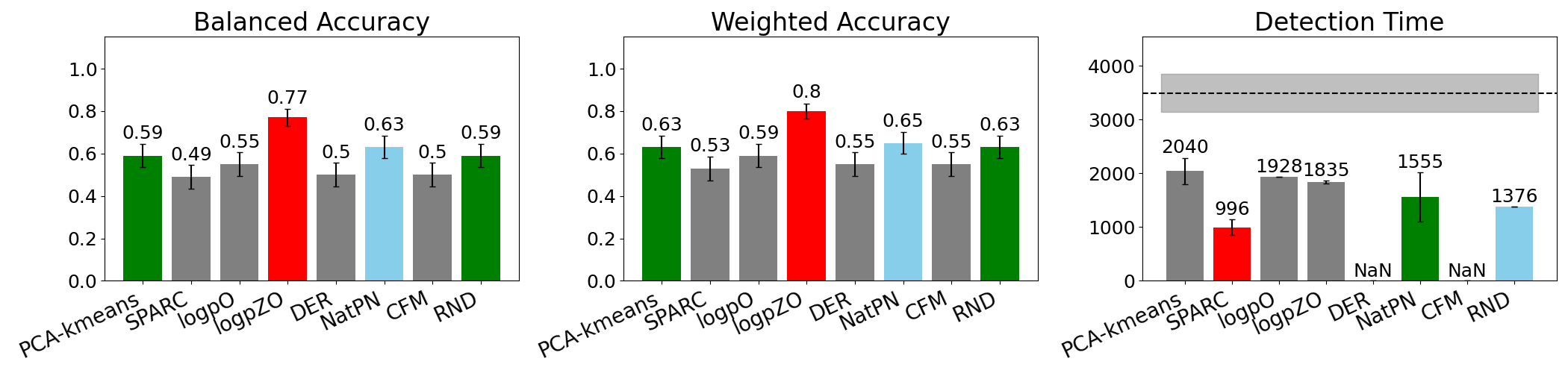}
        \vspace{-20pt}
        \subcaption{\makecell[l]{\textbf{CleanUpSpill} with DP:  (Setting-dependent band) OOD}}\label{fig:clean_spill}
    \end{minipage}
    \begin{minipage}[b]{0.495\linewidth}
        \includegraphics[width=\linewidth]{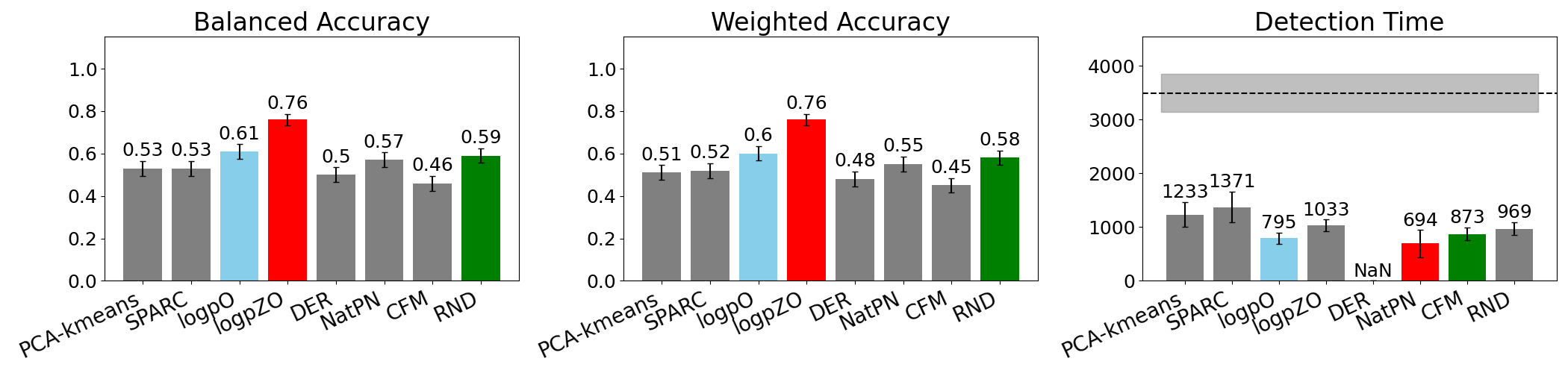}
        \vspace{-20pt}
        \subcaption{\makecell[l]{\textbf{CleanUpSpill} with DP:  (ID-only band) OOD}}
    \end{minipage}
    \caption{\small Quantitative results for the robot hardware experiments across two tasks with policies trained using FM and DP. We consider two different ways to compute the CP band: ``setting-dependent'' using successful trajectories from each OOD/ID environment and ``ID-only'' using only the trajectories from the ID environment. For balanced accuracy and weighted accuracy, higher is better and for detection time, lower is better. Additional metrics are reported in \cref{fig:real_metrics} and \cref{fig:real_metrics_Spill}. The figure layout is the same as \cref{fig:sim_metrics_abbrev} ({\color{red} best}, {\color{cyan} second}, {\textcolor{ForestGreen}{third}}), and `NaN' detection time indicates that no test rollout was detected as failed.
    Once again the learned approaches outperform the post-hoc methods. Note we do not present STAC here as it was slow to run on hardware in real-time. In the small sample size regime, logpZO remains robust in \textbf{combined accuracy}, achieving top-\rev{1 } performance in the highest number of cases (\rev{8}/12) and \rev{top-3 }performance in \rev{11}/12 cases. RND \rev{underperforms by never reaching top-1 performance, } yet \rev{it always achieves } top-3 performance. In contrast, the PCA-kmeans baseline reaches top-1 performance in \rev{4}/12 cases and top-3 performance in \rev{10}/12 cases. 
    In \textbf{detection time}, the \rev{post-hoc SPARC } method is the fastest in \rev{4}/6 cases, \rev{yet it never achieves top-1 performance. } PCA-kmeans is robust in speed as it attains top-3 performance in \rev{4}/6 cases. logpZO still remains practical with detection times well below the average success trajectory completion time.
    }
    \label{fig:real_metrics_abbrev}
    \vspace{-10pt}
\end{figure*}

\subsection{Sequential Threshold Design with Conformal Prediction}\label{sec:threshold}
We design a time-varying threshold $\eta_t$ such that a failure is flagged when $D_M(A_t, O_t;\theta)$ exceeds~$\eta_t$. To do so, we leverage functional CP \citep{diquigiovanni2024importance}, a framework that wraps around a time series of any scalar score $D_M(A_t,O_t;\theta)$ (higher indicates failure) and yields a distribution-free prediction band $C_{\alpha}$ with user-specified significance level $\alpha \in (0,1)$. Under mild conditions \citep{vovk2005algorithmic,xu2023conformal,xu2023sequential}, $C_{\alpha}$ contains any ID score $D_M(A_t,O_t;\theta)$ with probability of at least $1-\alpha$ for the entire duration of the rollout. If $D_M(A_t,O_t;\theta)\notin C_{\alpha}$, we can confidently reject that $(A_t,O_t)$ is ID. 

For sequential failure detection, we build $C_{\alpha}$ as a one-sided time-varying CP band. The band is one-sided as we are only concerned with high values of the scalar score $D_M(A_t,O_t;\theta)$, which indicate the trajectory is OOD (i.e., a failure). Given $N$ successful rollouts as the calibration data, we obtain scalar scores $\mathcal{D}_{cal}=\{D_M(A_t^i,O_t^i;\theta): i=1,\ldots,N \text{ and } t=1,H',\ldots,T\}$. The CP band is a set of intervals $C_{\alpha}=\{[\text{lower}_t,\text{upper}_t]: t=1,H',\ldots,T\}$, where $\text{lower}_t~\equiv~\min(\mathcal{D}_{cal})$ since the band is one-sided. To obtain the upper bound, we follow \citep{diquigiovanni2024importance}, computing the time-varying mean $\mu_t$ and band width $h_t$, so that $\text{upper}_t=\mu_t+h_t$. Further details of upper bound construction are in \cref{appendix:CP_band}.
Theoretically, for a new successful rollout $\tau_T=(O_0,A_0,\ldots,O_T,A_T)$, with probability at least $1-\alpha$, the score $D_M(A_t,O_t;\theta) \in [\text{lower}_t,\text{upper}_t]$ for all $t=1,H',\ldots,T$. By defining the threshold $\eta_t=\text{upper}_t$ and setting failures to one,
the decision rule $\mathbbm{1}(D_M(A_t,O_t;\theta) > \eta_t)$ controls the false positive rate (successes marked as failures) at level $\alpha$.

\section{Experiments}
\label{sec:experiments}

We test our two-stage failure detection framework in both simulation and on robot hardware. Our experiments span multiple environments, each presenting unique challenges in terms of types of tasks and distribution shifts. We empirically investigate an extensive set of both learned and post-hoc scalar scores (see \cref{tab:score_overview}) within our FAIL-Detect framework (see results in \cref{sec:results}).
We refer to \cref{appendix:expr} for more details on policy training, the CP band calibration procedure, and the learned scalar score architectures. 

\paragraph{Tasks} In simulation, we consider the \textbf{Square}, \textbf{Transport}, \textbf{Can}, and \textbf{Toolhang} tasks from the open-source Robomimic benchmark\footnote{We omit the \textbf{Lift} task as both FM and DP policies achieve 100\% success.}~\citep{mandlekarmatters}. 
In the robot hardware experiments, we consider two tasks on a bimanual Franka Emika Panda robot station that are significantly more challenging: \textbf{FoldRedTowel} and \textbf{CleanUpSpill} (see \cref{fig:real_demo}). 
We construct OOD settings for each task. In simulation, we adjust the third-person camera \SI{10}{\centi\meter} upwards at the first time step after $t=50$ to simulate a camera bump mid-rollout\footnote{We use $t=15$ for \textbf{Can}, which has the shortest task completion time.}. For the on-robot \textbf{FoldRedTowel} task, we disturb the task after the first fold (challenging ID scenario) and create an OOD initial condition by crumpling the towel (seen in less than $\sim$15\% of the data) and adding a never before seen distractor (blue spatula). For the \textbf{CleanUpSpill} task, we create an OOD initial condition by changing the towel to a novel green towel (see \cref{fig:real_OOD}).

\paragraph{Baselines}
We baseline FAIL-Detect against STAC~\citep{agia2024unpacking} and PCA-kmeans \citep{liu2024multitask} as SOTA approaches in success-based failure detection for generative imitation learning policies. 
STAC operates by generating batches (e.g., 256) of predicted actions at each time step. It then computes the statistical distance (e.g., maximum mean distance (MMD)) between temporally overlapping regions of two consecutive predictions, where the MMD is approximated by batch elements. Intuitively, the MMD measures the ``surprise" in the predictions over the rollout and subsequently, STAC makes a detection using CP. Note that instead of computing a CP band for a temporal sequence, STAC computes a single threshold based on empirical quantiles of the cumulative divergence in a calibration set. We reproduce the method and adopt hyperparameters used in their push-T example, where we generate a batch of 256 action predictions per time step. We did not employ the VLM component of the STAC failure detector to remain as real-time feasible as possible. Due to the long STAC inference time (even after parallelization) and resulting high system latency, we omit its comparison on the two robot hardware tasks.
In our second baseline, \citet{liu2024multitask} tackle failure detection by training a failure classifier, which requires the collection of failure training data. However, this approach is not applicable to our setup as we assume access to only successful human demonstrations for training and successful rollouts for calibration. Instead, we incorporate their proposed OOD detection method as a post-hoc scalar score in the first stage of FAIL-Detect to construct a fair baseline. We use the performant time-varying CP band to obtain thresholds in the second stage. The method measures the distance of a new observation $O_{t'}$ at test time index $t'$ from the set of training data $\{O_t\}_{t \geq 0}$, which consist of visual encoded features jointly trained with the policy on the demonstration data. PCA-kmeans first uses PCA to embed the training features and then applies $K$-means clustering to the embedded data to obtain $K=64$ centroids. After embedding $O_{t'}$ using the same principal components, the method computes the smallest Euclidean distance between the embedding and the $K$ centroids. This distance serves as the OOD metric (higher values indicate greater OOD). 
%
We omit comparison against ensembles~\citep{lakshminarayanan2017simple}, a popular OOD detection technique, due to RND having shown improved performance over ensembles in prior work~\citep{Ciosek2020Conservative} and their prohibitively high computational cost.

\paragraph{Evaluation Protocol} \label{subsec:evaluation_protocol}
To quantify failure detection performance, we denote failed rollouts as one and successful rollouts as zero. We then adopt the following standard metrics: 
\textbf{(1)} true positive rate (TPR), 
\textbf{(2)} true negative rate (TNR), 
\textbf{(3)} balanced accuracy = (TPR + TNR) / 2,
\textbf{(4)} weighted accuracy = $\beta\cdot$TPR + $(1-\beta)\cdot$ TNR for $\beta=\frac{\# \text{Successful rollouts}}{\# \text{Rollouts}}$, and
\textbf{(5)} detection time = $\mathbb{E}_{(A_t,O_t)\sim \text{test rollouts}}[\arg\min_{t=1,H',\ldots,T} \mathbbm{1}(D_M(A_t,O_t;\theta)>\eta_t)]$, which computes the average failure detection time from the start of the rollout. The balanced accuracy metric equally represents classes in an imbalanced dataset (e.g., few successful rollouts in an OOD setting). Weighted accuracy represents how well a method matches the true success / failure distribution.
Due to the high human time cost of performing real-robot rollouts, we evaluate FAIL-Detect and the baselines on significantly fewer rollouts in the robot hardware tasks (i.e., 50 rollouts) compared to the simulation tasks (i.e., 2000 rollouts).

\begin{figure*}[!t]
    \centering
    \begin{minipage}{0.19\linewidth}
        \includegraphics[width=\linewidth]{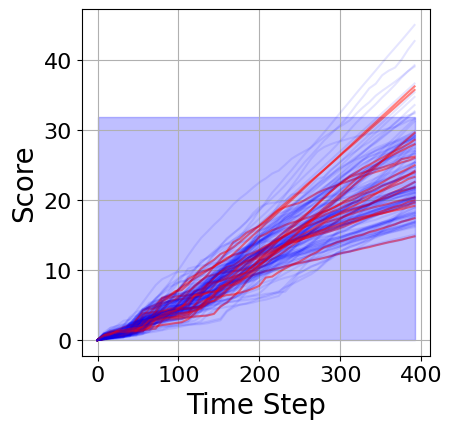}
        \subcaption{STAC}
    \end{minipage}
    \begin{minipage}{0.195\linewidth}
        \includegraphics[width=\linewidth]{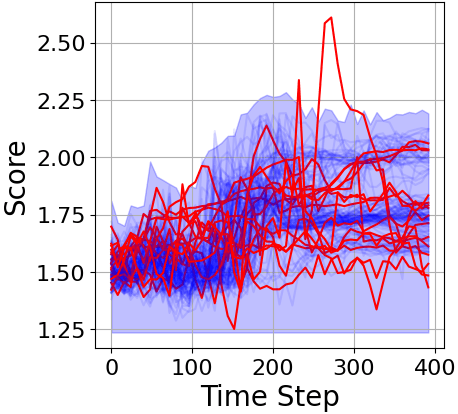}
        \subcaption{PCA-kmeans}
    \end{minipage}
    \begin{minipage}{0.195\linewidth}
        \includegraphics[width=\linewidth]{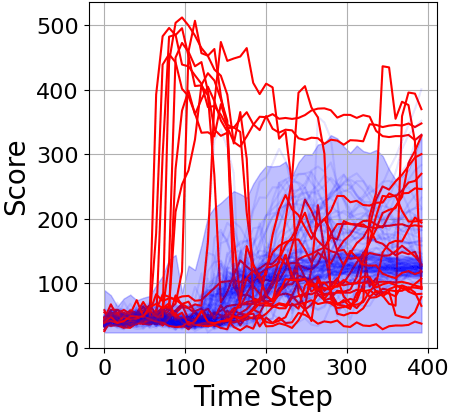}
        \subcaption{logpZO}
    \end{minipage}
    \begin{minipage}{0.192\linewidth}
        \includegraphics[width=\linewidth]{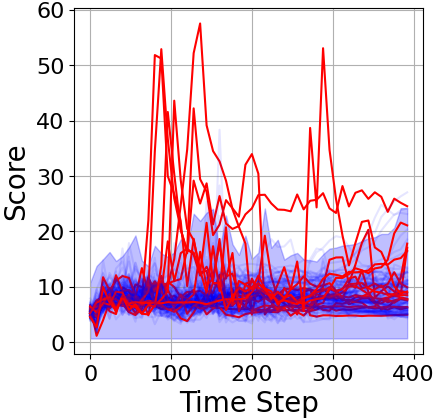}
        \subcaption{NatPN}
    \end{minipage}
    \begin{minipage}{0.185\linewidth}
        \includegraphics[width=\linewidth]{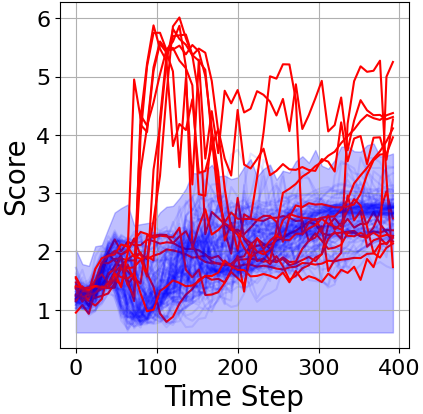}
        \subcaption{RND}
    \end{minipage}
    \caption{\small  Qualitative results of failure detection scores overlaid with CP bands. The curves are colored by the ground truth success/failure status of the rollout (failure = {\color{red}red} and success = {\color{blue}blue}). We show 150 test rollouts on \textbf{Square ID} across post-hoc baselines (STAC, PCA-kmeans) and learned FAIL-Detect methods (logpZO, NatPN, RND). We use the constant CP threshold for STAC as per~\cite{agia2024unpacking}. 
    Note that post-hoc baseline methods mark most trajectories as successes due to the poor failure/success separation. In comparison, learned metrics have tight CP bands and higher failure/success separation.}
    \label{fig:score_visual}
    \vspace{-10pt}
\end{figure*}

\section{Results}\label{sec:results}

We present our experimental findings addressing the following research questions:
\begin{enumerate}
    \item[A.] How performant is failure detection without failure data?
    \item[B.] What is the impact of learned vs. post-hoc scores on failure detection?
    \item[C.] Do failure detections align with human intuition?
\end{enumerate}

\subsection{How performant is failure detection without failure data?}
A key question we consider is whether failure detection is possible and performant without enumerating all possible failure scenarios, which is practically infeasible. We conduct extensive experiments across simulation and robot hardware tasks to answer this question. We evaluate balanced accuracy, weighted accuracy, and detection time to assess whether failures can be identified reliably and quickly. 

\textbf{FAIL-Detect achieves high accuracy with fast detection.} Our two-stage framework demonstrates strong performance across both accuracy metrics and detection speed. For example, the average \textit{best} balanced accuracy across FAIL-Detect's score candidates is $\sim\rev{78}\%$ in simulation (\cref{fig:sim_metrics_abbrev}) and $\sim\rev{72}\%$ on the robot hardware tasks (\cref{fig:real_metrics_abbrev}). This performance shows the capacity of failure-free failure detection methods to robustly identify failures across many scenarios. Notably, FAIL-Detect maintains viable detection time across various score designs, with average \textit{best} detection time faster than successful trajectory completion.

\subsection{What is the impact of learned vs. post-hoc scores on failure detection?}

\textbf{Learned scores outperform post-hoc scores.} 
Looking at performance across simulation and robot hardware tasks, we find that learned scalar scores hold an advantage over post-hoc scores in failure detection.
In simulation (\cref{fig:sim_metrics_abbrev}), logpZO and RND are the best two methods, achieving top-\rev{1 } performance in \rev{10}/16 and \rev{5}/16 cases, respectively. 
STAC is the best in the post-hoc category for \mbox{top-\rev{1}} accuracy in \rev{3}/16 cases, yet PCA-kmeans is never the best. Overall, there is a large performance gap between the learned and post-hoc methods, especially in terms of the best overall accuracy. We did notice that post-hoc methods perform better in the OOD cases than in ID scenarios. We hypothesize this may be due to a clearer distinction between successful ID trajectories and failed OOD trajectories, for example, for SPARC if the OOD trajectory exhibits significant jitter.

In terms of detection time, logpZO is the most efficient, achieving the fastest time in \rev{3}/8 cases, \rev{while PCA-kmeans does so in only 1/8 cases}. 
Notably, STAC's detection time consistently exceeds practical limits, surpassing the average success trajectory time. 

\begin{figure*}[!t]
    \centering
    \begin{minipage}[b]{0.245\textwidth}
        \includegraphics[width=0.49\linewidth]{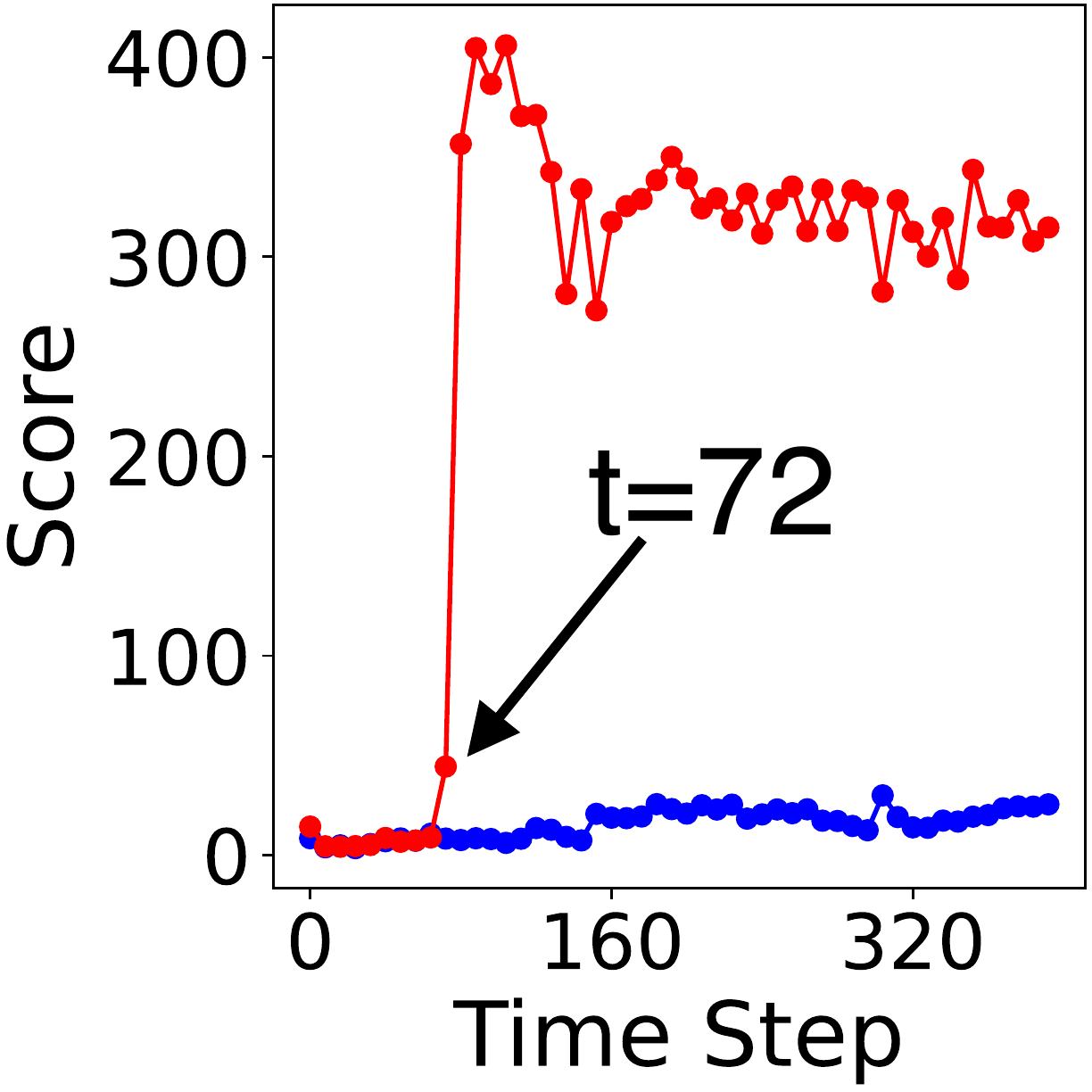}
        \includegraphics[width=0.49\linewidth]{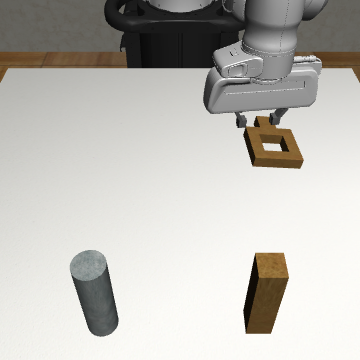}
        \subcaption{Square}
        \label{fig:square}
    \end{minipage}
    \begin{minipage}[b]{0.245\textwidth}
        \includegraphics[width=0.49\linewidth]{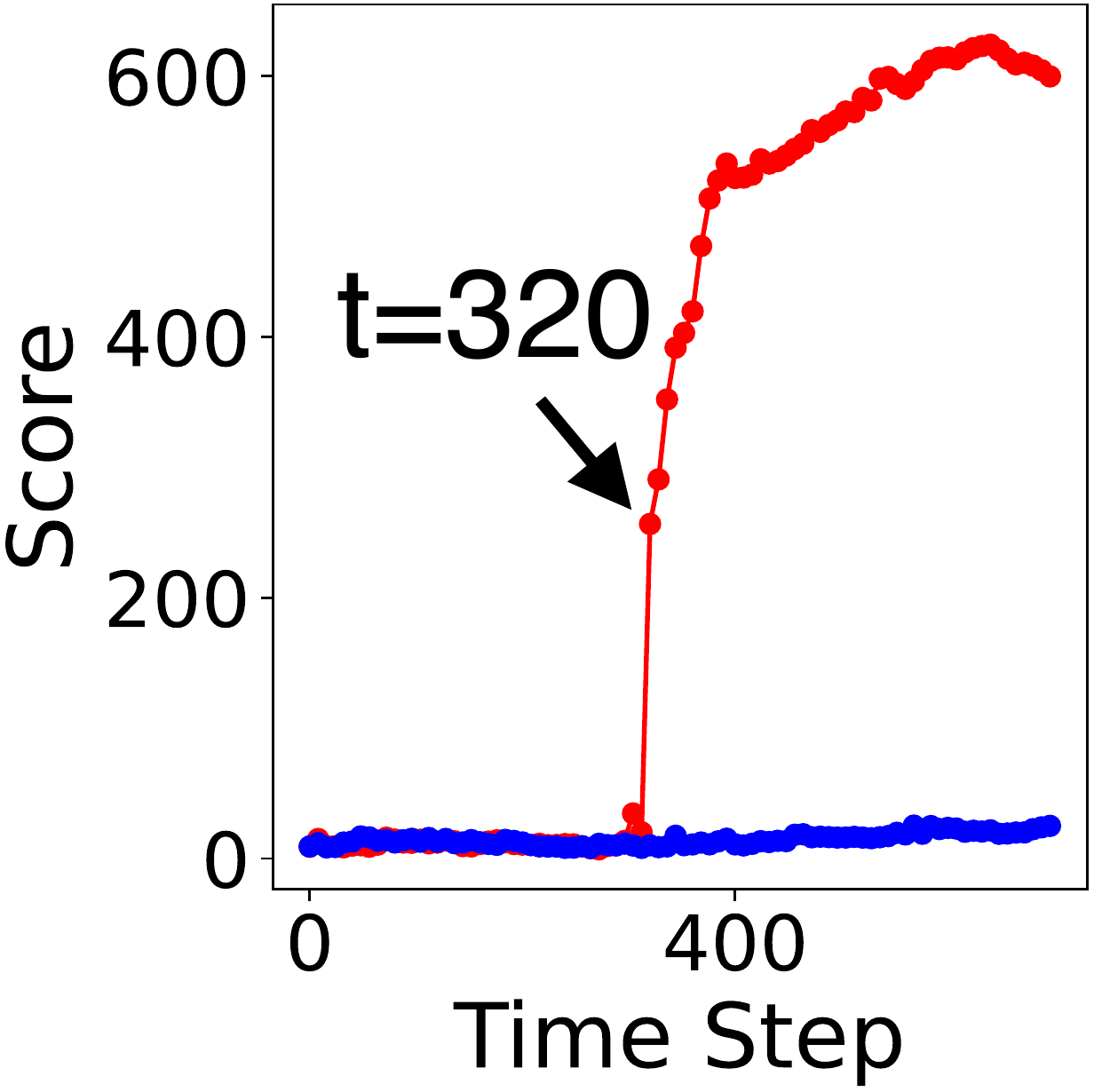}
        \includegraphics[width=0.49\linewidth]{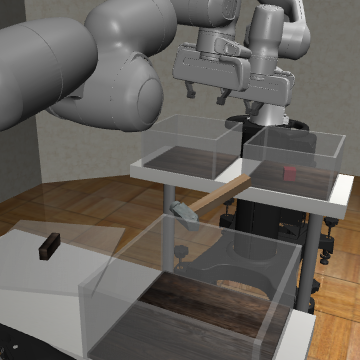}
        \subcaption{Transport}
        \label{fig:transport}
    \end{minipage}
    \begin{minipage}[b]{0.245\textwidth}
        \includegraphics[width=0.49\linewidth]{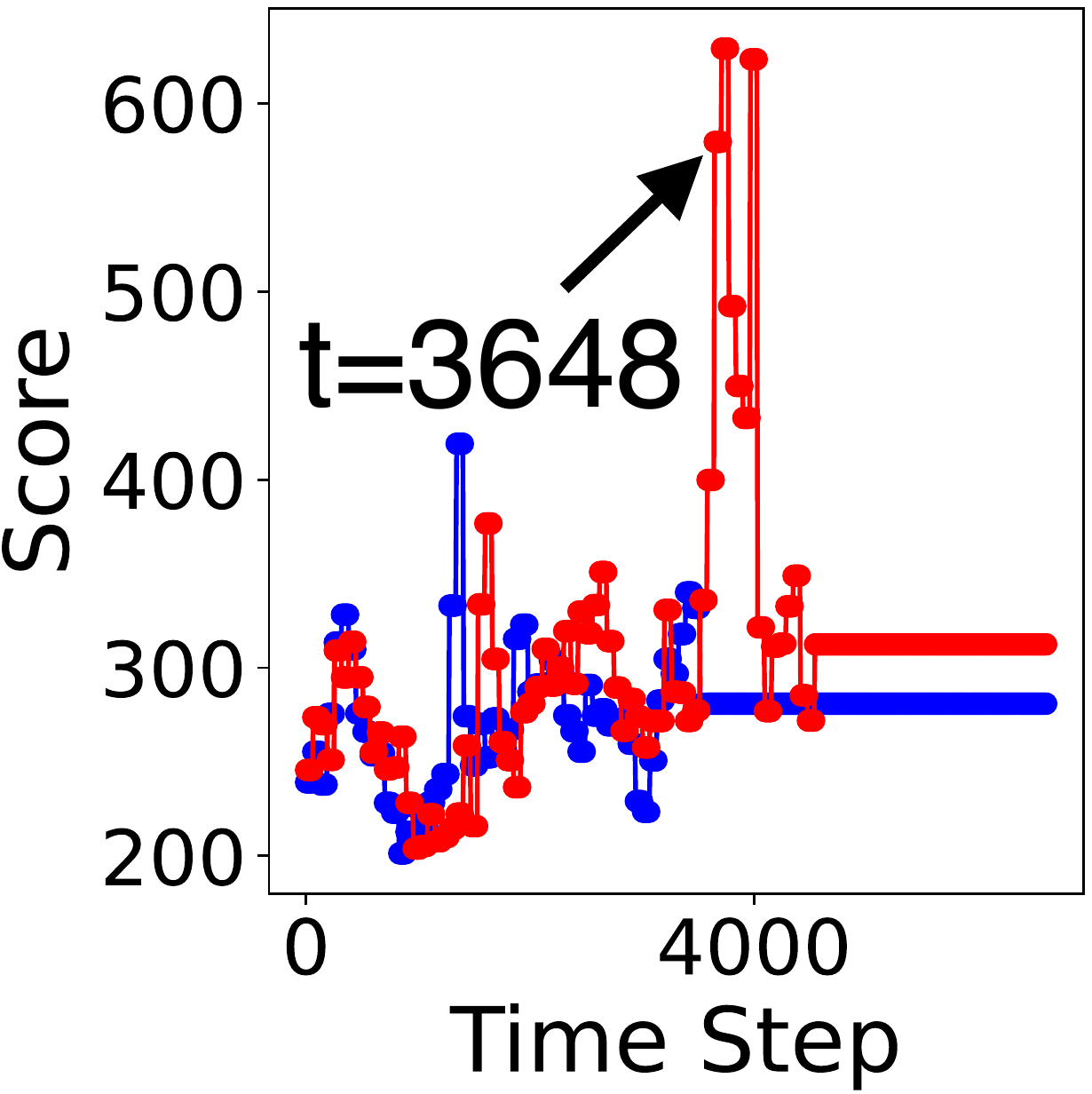}
        \includegraphics[width=0.49\linewidth]{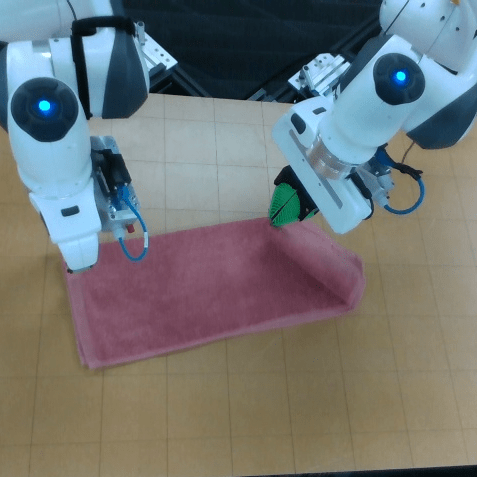}
        \subcaption{FoldRedTowel ID + Disturb}
        \label{fig:towel1}
    \end{minipage}
    \begin{minipage}[b]{0.245\textwidth}
        \includegraphics[width=0.49\linewidth]{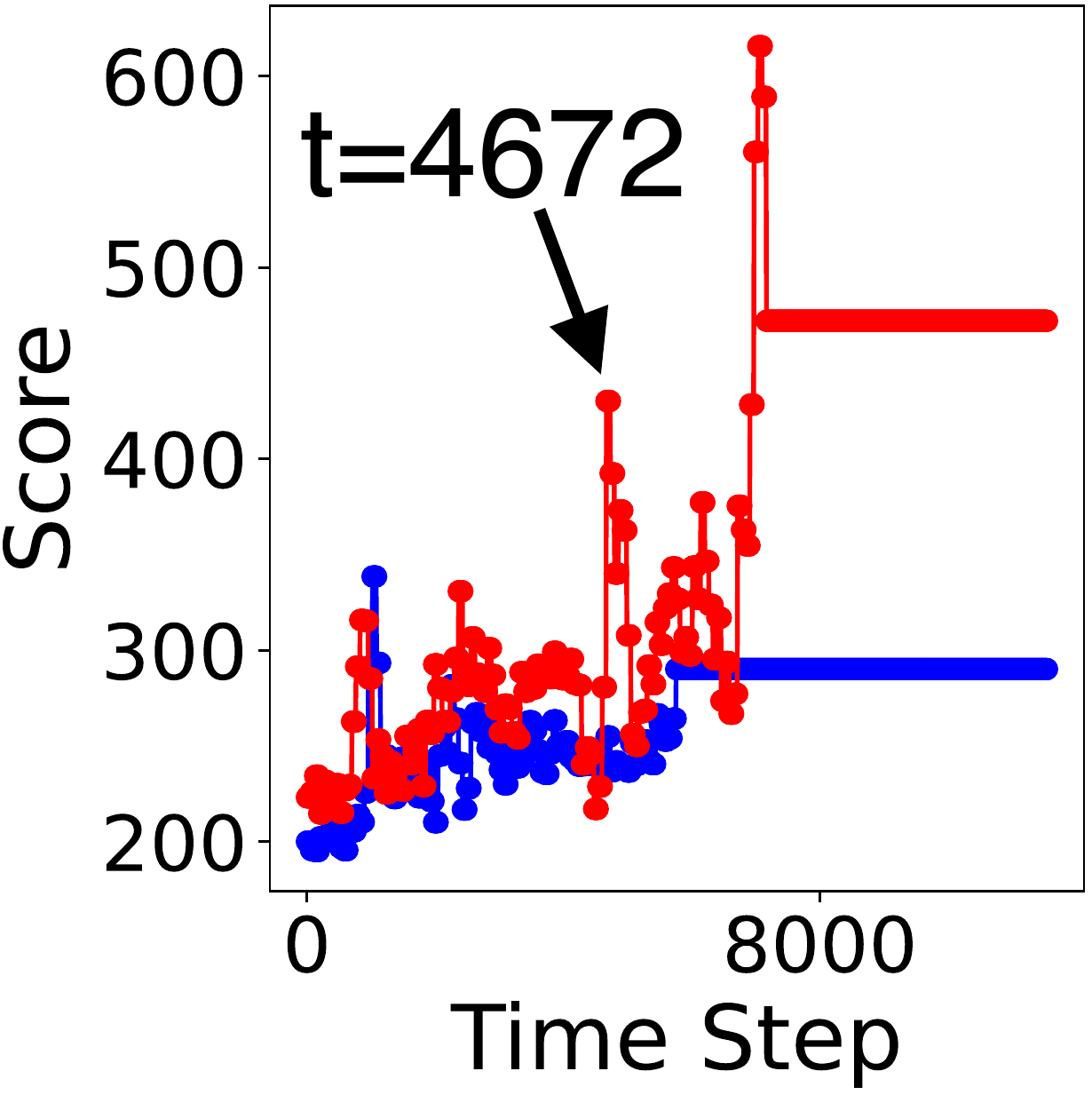}
        \includegraphics[width=0.49\linewidth]{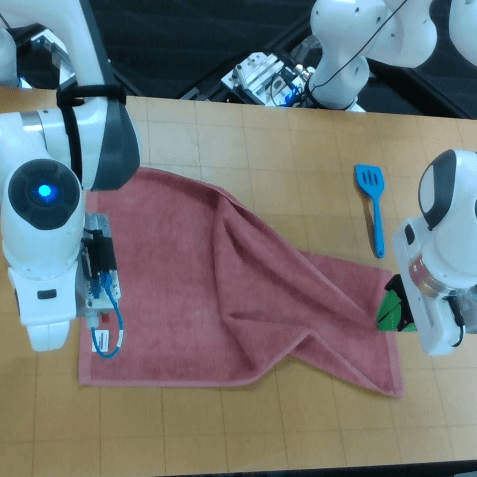}
        \subcaption{FoldRedTowel OOD}
        \label{fig:towel2}
    \end{minipage}
    \caption{\small Physical interpretation of logpZO, the most successful and robust learned score method. Failed trajectory scores are in {\color{red}red} and successful ones are in {\color{blue}blue}. Each figure shows the failure detection time and the corresponding camera view. \textbf{(Simulation)} In \cref{fig:square}, failure is flagged when the square slips from the gripper. In \cref{fig:transport}, failure is detected when both arms drop the hammer. \textbf{(On-robot)} In \cref{fig:towel1}, failure is alerted as the second fold attempt fails. In \cref{fig:towel2}, failure is detected as the left robot arm fails to complete the first fold.}
    \label{fig:physical_meaningful}
    \vspace{-10pt}
\end{figure*}

For the robot hardware experiments (\cref{fig:real_metrics_abbrev}), with a much lower rollout data regime for calibration than in simulation (see \cref{tab:eval_protocol_params}) and a wider diversity of behaviors and observations, logpZO remains robust as the best method, reaching \mbox{top-\rev{1 }}highest balanced accuracy and weighted accuracy in \rev{8}/12 scenarios. The PCA-kmeans baseline \rev{is second best  with \rev{4}/12 top-1 ranking}. \rev{RND underperforms by never achieving top-1 performance, yet it always ranks among the top-3 best methods. }
Additionally, most methods, including logpZO and RND, maintain practical detection times well below the average successful trajectory time. \rev{SPARC } is on average the fastest as it attains top-\rev{1 } performance in \rev{4}/6 cases, \rev{but it exhibits poor accuracy.}

Overall, across all experiments, we find our novel learned logpZO score within the FAIL-Detect framework to be most consistent in performance. The post-hoc methods are often at the extremes of performance (either doing well or poorly) depending on the particular setting.

\textbf{Qualitative score trends.} Visualization of the detection scores (\cref{fig:score_visual,fig:score_visual_real}) confirms that learned methods are more discriminative with better score separation between successful and failed trajectories compared to post-hoc approaches. 
STAC also suffers from a single calibration threshold that is time invariant.
In \cref{sec:ablation}, we present comprehensive ablation studies examining
performance sensitivity to CP significance level $\alpha$ (\cref{fig:alpha_ablate}).

\textbf{Computational advantage.} 
Some post-hoc methods require sampling from the stochastic policy repeatedly to achieve a performant failure score. For example, STAC requires generating 256 action predictions per time step. Although the computational efficiency could be improved by generating fewer predictions, this compromises its statistical reliability. On the other hand, our learned scores offer significantly faster inference speeds compared to STAC.
For instance, testing on an A6000 GPU with 50 rollouts, logpZO score computation takes \SI{0.04}{\second} (\textbf{Square}) and \SI{0.033}{\second} (\textbf{Transport}) per time step, while STAC requires \SI{1.45}{\second} for both tasks, amounting to a 36-44 times slowdown. 

\subsection{Do failure detections align with human intuition?} 

FAIL-Detect's alerts demonstrate strong correlation with observable failure indications in the environment (\cref{fig:physical_meaningful}). When scores exceed the decision threshold, these moments often align with meaningful changes in the physical state of the task.
In simulation environments, the detection scores capture distinct failure patterns with high precision. For instance, for the \textbf{Square} task, abrupt increases in scores coincide with the moment the gripper loses its hold on the square. Similarly, in the \textbf{Transport} task, score spikes identify the instant when the hammer slips during inter-arm transfer. Real-world applications demonstrate similarly compelling results: the system effectively detects both human-induced disruptions leading to an incomplete second towel fold (\textbf{FoldRedTowel ID + Disturb}) and OOD initial conditions resulting in an improper first towel fold (\textbf{FoldRedTowel OOD}).

This correspondence between score spikes and physical events is encouraging for FAIL-Detect's capacity to capture task-relevant subtlety. The framework successfully translates complex environmental changes into quantifiable metrics, with score variations serving as reasonable indicators of failure events.
Moreover, this correlation offers valuable diagnostic capabilities for potential policy improvement. Instead of requiring an exhaustive a priori enumeration of potential failure modes, which is an inherently challenging endeavor, our approach enables potential targeted analysis of observed failures. By examining executions within temporal windows surrounding a failure detection, one can efficiently identify failure types for subsequent analysis.

\textbf{Environment-dependent thresholds aid performance on robot hardware tasks.} 
For simulation tasks (\cref{fig:sim_metrics_abbrev}), ID-only bands computed on successful ID rollouts work well for failure identification in both the ID and the OOD camera bump scenarios. This setting is practically preferable as it does not require collecting successful rollouts in each new environment to calibrate the failure detection threshold. 
However, on-robot tasks with OOD initial conditions (\cref{fig:real_metrics_abbrev,fig:real_metrics,fig:real_metrics_Spill}), most methods show degraded performance with ID-only bands, at times yielding low or close to zero TNR due to most rollouts being marked as failure. This over-conservative behavior likely occurs because OOD successful trajectories are less performant. 
We find on-robot policies to be more sensitive to the environmental changes we employed than the simulation-based policies in response to the camera bump. 
Even when tasks are completed successfully in these challenging conditions, trajectories often exhibit poor quality (e.g., slow execution, jitter) and higher (worse) scores compared to ID successful rollouts, suggesting they should potentially be classified as failures. 

\section{Limitations}
With FAIL-Detect, we demonstrate the promising potential of a failure detection method without access to failure data. The proposed two-staged method does have several limitations, which offer possible avenues for future work. We observed that, at times, the learned scores focus on simpler robot state information (e.g., gripper closed or open) over the higher-dimensional visual features. Finding score architectures that maximally make use of the visual information could improve failure detection performance. Meanwhile, we acknowledge the existence of false positives, especially in OOD settings where the trajectories degrade in performance. Our proposed setting-dependent CP band is meant to account for these environmental and behavioral shifts. However, more advanced techniques such as adaptively adjusting the CP significance level during inference could be helpful. Additionally, our score candidates do not consider long sequences of temporal data. Although the CP band is temporally-aware, having the learned scores distill temporal patterns from historical time series data may lead to better failures predictions. Lastly, the timely detection of failures may be further improved by considering multimodal sensory information such as sound or tactile.

\section{Conclusion}
We present a modular two-stage uncertainty-aware runtime failure detection approach for generative imitation learning in manipulation tasks. Our method combines a learned scalar score and time-varying conformal prediction to accurately and efficiently identify failures while providing statistical guarantees. 
Through extensive experiments on diverse tasks, we show that the novel logpZO score consistently achieves the highest performance among the considered candidates. Crucially, we demonstrate that unseen failures can be effectively detected without access to failure data, which may be difficult to collect to cover all possible failure scenarios. Our results highlight the potential of our method to enhance the safety and reliability of robotic systems in real-world applications.


\section*{Acknowledgement}
This work was primarily done during an internship at Toyota Research Institute (TRI). We thank Vitor Guizilini, Benjamin Burchfiel, and David Snyder for the helpful discussions and feedback. We thank the robot teacher team at the TRI headquarters for the data collection, specifically Derick Seale and Donovon Jackson.

\bibliographystyle{plainnat}
\bibliography{references}

\clearpage
\onecolumn
\appendix

\subsection{Proposed logpZO score network}\label{appendix:logpZO}
We explain how the proposed novel logpZO works step by step:
\begin{itemize}
    \item \textbf{Step 1:} Fit a flow matching model $f_{\theta}$ between observations $\{O_t\}$ (i.e., image embeddings and proprioception) and latent noise $\{Z\}\sim \mathcal{N}(0,I)$, so that for $s\in [0,1]$:
    \begin{align*}
        f_{\theta}(O_t[s], s) & \approx Z - O_t, \\
        O_t[s] & = O_t + s(Z-O_t).
    \end{align*}

    \item \textbf{Step 2:} Given a new observation $O_{t'}$ at time step $t'$, perform one-step prediction to obtain latent noise estimate:
    \[
    Z_{O_{t'}} = O_{t'} + f_{\theta}(O_{t'}, 0).
    \]
    When $O_{t'}$ is in-distribution, by the flow matching formulation, $Z_{O_{t'}}$ is close to samples drawn from $\mathcal{N}(0, I)$.

    \item \textbf{Step 3:} Compute density of latent noise (up to a constant) using squared norm:
    \[
    \log p(Z_{O_{t'}}) \propto -\|Z_{O_{t'}}\|_2^2
    \]
    High norm values of $\|Z_{O_{t'}}\|_2^2$ indicate lower likelihood, which is caused by anomalous observations $O_{t'}$. We thus use $\|Z_{O_{t'}}\|_2^2$ as the logpZO score.
\end{itemize}

\subsection{CP band construction}\label{appendix:CP_band}
Following \citep{diquigiovanni2024importance}, we split the set of calibration scores $\mathcal{D}_{cal}$ into two disjoint parts $\mathcal{D}_{cal_A}$ and $\mathcal{D}_{cal_B}$ with sizes $N_1$ and $N_2$. We first compute the mean successful trajectory $\mu_t~=~N_1^{-1} \sum_{i=1}^{N_1} D_M(A^i_t, O^i_t; \theta)$ for $t=1,\ldots,T$ on $\mathcal{D}_{cal_A}$. Then, for $j=1,\ldots,N_2$, we compute $D_j=\max(\{(\mu_t-D_M(A^j_t, O^j_t; \theta))/s_{cal_A}(t)\}_{t=1}^T)$, which is the max deviation over rollout length from the mean prediction to the scalar score. The function $s_{cal_A}(t)$ is called a ``modulation'' function that depends on the dataset $\mathcal{D}_{cal_A}$. In our experiment, we consider either
\begin{align}
    s_{cal_A}(t)&=1/T \label{eq:v1} \\
    s_{cal_A}(t)&=\max_{k\in \mathcal{H}} |D_M(A^k_t, O^k_t,\theta)-\mu_t| \label{eq:v2},
\end{align}
where $\mathcal{H}=[N_1]$ if $(N_1+1)(1-\alpha)>N_1$, otherwise $\mathcal{H}=\{k \in [N_1]: \max_{t\in [T]} |D_M(A^k_t, O^k_t,\theta)-\mu_t| \leq \gamma\}$ for $\gamma=(1-\alpha)\text{-quantile of }\{\max_{t\in [T]} |D_M(A^m_t, O^m_t,\theta)-\mu_t|\}_{m=1}^{N_1}.$
Intuitively, \cref{eq:v2} adapts the width of prediction bands based on the non-extreme behaviors of the functional data. It does so by minimizing the influence of outliers whose maximum absolute residuals lie within the upper $\alpha$ quantile of all maximum values.
Additionally, note that the $\max$ is taken because the CP band is intended to reflect the entire trajectory. We define $S=\{D_j, j=1,\ldots, N_2\}$ as the collection of such max deviations. The band width $h$ is finally computed as the $(1-\alpha)$-quantile of $S$ and the upper bound is $\text{upper}_t=\mu_t+hs_{cal_A}(t)$. We pick $\alpha=0.05$ (or 95\% confidence interval) throughout experiments (see Table \cref{tab:eval_protocol_params} on hyperparameter choices). 

\subsection{Experimental Details}\label{appendix:expr}

\begin{table}[!b]
\centering
\caption{\small  Success rate of the flow policy on test data in each task-environment combination. These test data is used to test failure detection methods as well. In the hardware experiments, we mark some cells with $^*$ when the number of failures out of test rollouts is no greater than 5. In such cases, we shuffle the rollout indices and include all the failure ones in the test set, so that the failure detection metrics have higher statistical significance. Across the entire 50 rollouts, the true success rate of FM policy on FoldRedTowel ID is 0.96, and that of DP on CleanUpSpill ID is 0.82.}\label{tab:success_rate}

\begin{subtable}{\linewidth}
\centering
\subcaption{Simulation tasks}
\resizebox{\linewidth}{!}{
\begin{tabular}{ccccccccc}
\toprule
& Square ID & Square OOD & Transport ID & Transport OOD & Can ID & Can OOD & Toolhang ID & Toolhang OOD \\
FM Policy & 0.90 (1000 rollouts) & 0.63 (2000 rollouts) & 0.85 (1000 rollouts) &  0.63 (2000 rollouts) & 0.98 (1000 rollouts) &  0.84 (2000 rollouts) & 0.77 (1000 rollouts) &  0.53 (2000 rollouts) \\
DP & 0.93 (125 rollouts) & 0.63 (250 rollouts) & 0.84 (125 rollouts) &  0.76 (250 rollouts) & 0.98 (125 rollouts) &  0.95 (250 rollouts) & 0.82 (125 rollouts) &  0.54 (250 rollouts) \\
\bottomrule
\end{tabular}}
\end{subtable}

\begin{subtable}{\linewidth}
\centering
\subcaption{Robot hardware tasks}
\resizebox{\linewidth}{!}{
\begin{tabular}{cccccccc}
\toprule
& \makecell[c]{FoldRedTowel ID \\ via FM policy} & \makecell[c]{FoldRedTowel ID + Disturb \\ via FM policy}  & \makecell[c]{FoldRedTowel OOD \\ via FM policy} & \makecell[c]{CleanUpSpill ID \\ via DP} & \makecell[c]{CleanUpSpill OOD \\ via DP} & \makecell[c]{CleanUpSpill ID \\ via FM policy} & \makecell[c]{CleanUpSpill OOD \\ via FM policy} \\
Setting-dependent band & 0.9$^*$ (20 rollouts) & 0.75$^*$ (20 rollouts) & 0.60 (20 rollouts) & 0.70 (20 rollouts) & 0.45 (20 rollouts) & 0.90* (20 rollouts) & 0.90* (20 rollouts) \\
ID-only band & 0.9$^*$ (20 rollouts) & 0.90 (50 rollouts) & 0.58 (50 rollouts) & 0.70 (20 rollouts) & 0.52 (50 rollouts) & 0.90* (20 rollouts) & 0.76 (50 rollouts) \\
\bottomrule
\end{tabular}}
\end{subtable}

\end{table}

\subsubsection{Task descriptions} 
\begin{itemize}
    \item \textbf{(Simulation)} The tasks from the Robomimic benchmark~\citep{mandlekarmatters} are as follows. The \textbf{Square} task asks the robot to pick up a square nut and place it on a rod, which requires precision.
    The \textbf{Transport} task asks two robot arms to transfer a hammer from a closed container on a shelf to a target bin on another shelf, involving coordination between the robots. 
    The \textbf{Can} task asks the robot to place a coke can from a large bin into a smaller target bin, requiring greater precision than the \textbf{Square} task.
    The \textbf{Toolhang} task asks the robot to assemble a frame with several components, requiring the most dexterity and precision among the four tasks.
    \item \textbf{(Robot hardware)} In the \textbf{FoldRedTowel} task, the two robot arms must fold a red towel twice and push it to the table corner. In the \textbf{CleanUpSpill} task, one robot arm must lift a cup upright that has fallen and caused a spill, while the other robot arm must pick up a white towel and wipe the spills on the platform. Both tasks are long-horizon and require precision and coordination to manipulate deformable objects.
\end{itemize}

\subsubsection{Policy backbone and the calibration of CP bands}\label{sec:hyperparams}

\cref{tab:success_rate} shows success rate across the tasks. Meanwhile, See \cref{tab:eval_protocol_params} for hyperparameters regarding 
\begin{itemize}
    \item Dimension of actions $A_t$ and observations $O_t$ per task.
    \item Architecture details of the policy backbone $g$ and the choice of image encoder.
    \item Training specifics of $g$ (i.e., optimizer, learning rate and scheduler, and number of epochs).
    \item Number of successful rollouts used to calibrate the CP bands and the number of test rollouts. Note that on simulation tasks, we roll out DP fewer times because it requires significantly longer time (higher number of denoising steps) than FM policies to generate actions.
\end{itemize}

We further explain the design and training of the policy network $g$.
The underlying policy network $g$ is trained with flow matching~\citep{lipman2023flow} and/or diffusion models~\citep{ho2020denoising}. We follow the setup in \citep{chi2023diffusionpolicy} and use the same hyperparameters to train the policies. When using flow matching \citep{lipman2023flow} to train the policies, the only difference is that instead of optimizing with the diffusion loss, we change the objective to be a flow matching loss between $A_t|O_t$ and $Z$, the standard Gaussian.  
Image features are extracted using either a ResNet or a CLIP backbone trained jointly with $g$. These image features concatenated with robot state constitute observations $O_t$.

\subsubsection{Scalar failure detection scores} After learning the policy network $g$ with the ResNet encoder for camera images, we first obtain $\{(A_t,O_t)\}$ for each task using the same training demonstration data for policy network. For the post-hoc approach SPARC, it utilizes the arc length of the Fourier magnitude spectrum obtained from the trajectory. 
To learn and test the scalar scores, we adopt the following setup:

\begin{figure}[!t]
    \centering
    \begin{minipage}[b]{0.245\linewidth}
        \includegraphics[width=\linewidth]{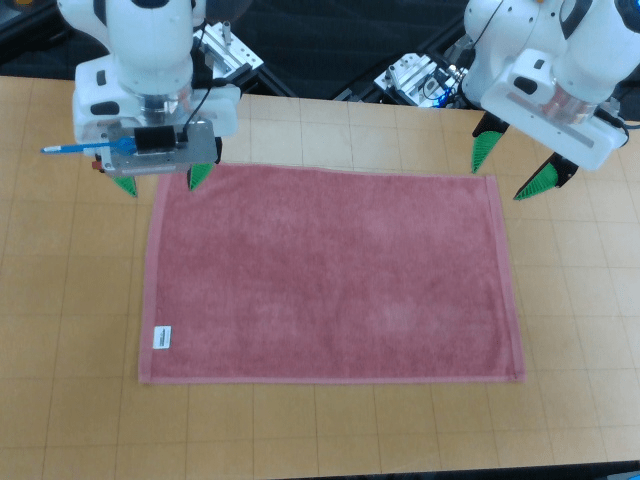}
        \subcaption{Initial condition}
    \end{minipage}
    \begin{minipage}[b]{0.245\linewidth}
        \includegraphics[width=\linewidth]{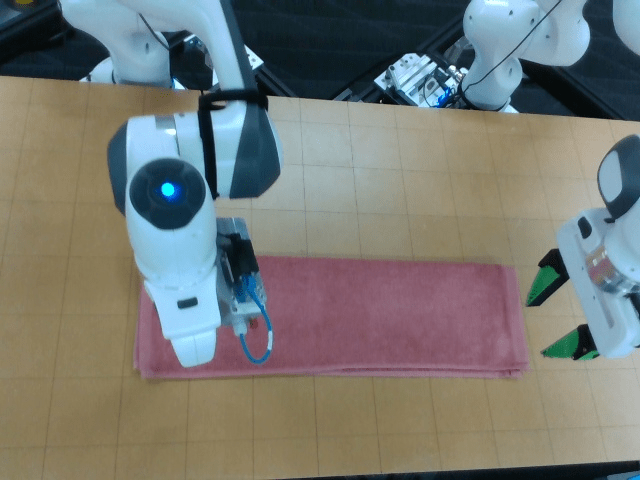}
        \subcaption{After 1st fold}
    \end{minipage}
    \begin{minipage}[b]{0.245\linewidth}
        \includegraphics[width=\linewidth]{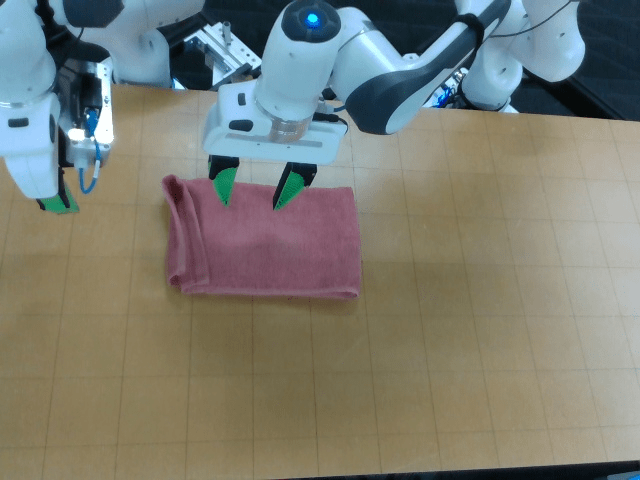}
        \subcaption{After 2nd fold}
    \end{minipage}\begin{minipage}[b]{0.245\linewidth}
        \includegraphics[width=\linewidth]{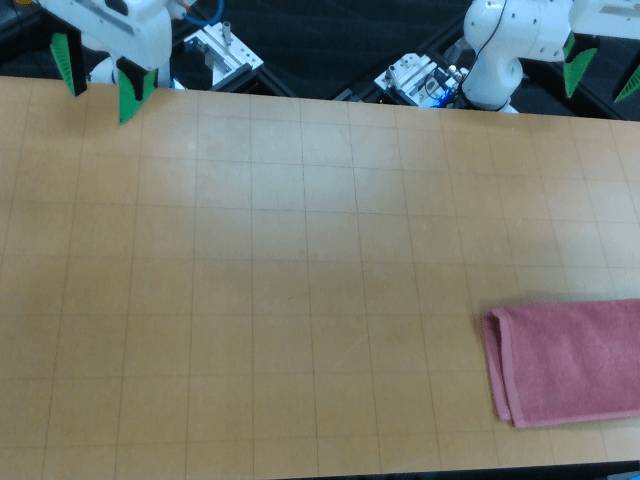}
        \subcaption{Final success}
    \end{minipage}
    \begin{minipage}[b]{0.245\linewidth}
        \includegraphics[width=\linewidth]{figs/Spill_ID_start.png}
        \subcaption{Initial condition}
    \end{minipage}
    \begin{minipage}[b]{0.245\linewidth}
        \includegraphics[width=\linewidth]{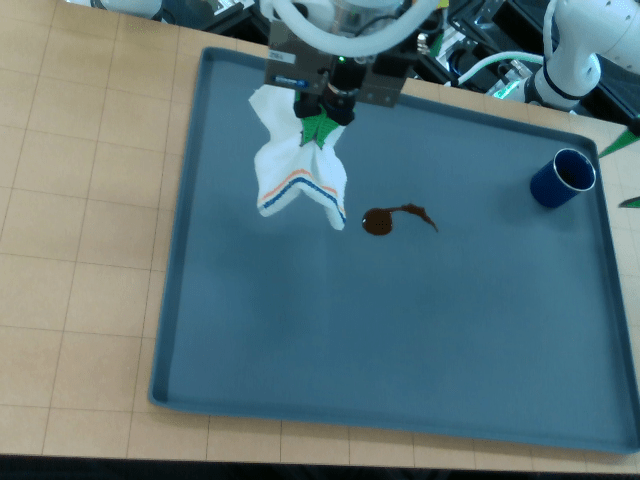}
        \subcaption{About to wipe}
    \end{minipage}
    \begin{minipage}[b]{0.245\linewidth}
        \includegraphics[width=\linewidth]{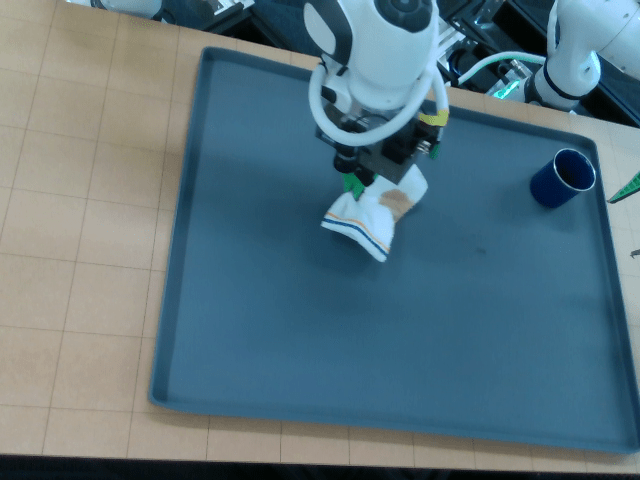}
        \subcaption{Wiping}
    \end{minipage}\begin{minipage}[b]{0.245\linewidth}
        \includegraphics[width=\linewidth]{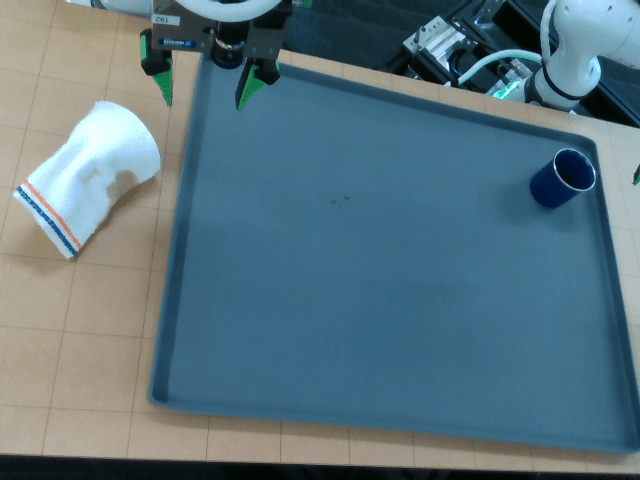}
        \subcaption{Final success}
    \end{minipage}
    \caption{\small  The on-robot experimental settings. \textbf{(Top row)} \textbf{FoldRedTowel}: starting with a flat towel, the two arms need to first fold the towel along the short side, and then the right arm needs to perform the second fold along the long side. Finally, the towel needs to be pushed to the bottom right corner to be considered a success. \textbf{(Bottom row)} \textbf{CleanUpSpill}: starting with spills caused by a fallen cup on the platform, the right arm must first lift the cup to an upright position, while the left arm must pick up a towel and wipe the spills. To achieve success, the spills must be completely cleaned, and the towel must be returned to its original position.}
    \label{fig:real_demo}
\end{figure}

\begin{enumerate}
    \item CFM: We use a 4x smaller network with identical architecture as the policy network. It is unconditional and takes in observations $O_t$ as inputs. We train for 200 epochs with a batch size of 128, using the Adam optimizer \citep{Kingma2014adam} with a constant 1e-4 learning rate.
    \item lopO and logpZO: We let the flow network (taking $O_t$ as inputs) has the same architecture as the policy network. On simulation, we let the flow network to be 4x smaller than the policy network with identical architecture and on real data, we keep identical model sizes between the two. On simulation (resp. real data), we train for 500 (resp. 2000) epochs with a batch size of 128 (resp. 512), using the Adam optimizer with a constant 1e-4 learning rate. For a new observation $O_{t'}$, its density $\log p(O_{t'})$ is obtained via the instantaneous change-of-variable formula~\citep{chen2018neural}.
    \item DER: The network to parametrize the Normal-Inverse Wishart parameters has the same architecture as the policy network but is 4x smaller with identical architecture. It takes in $O_t$ as inputs. We train for 200 epochs with a batch size of 128, using the Adam optimizer with a constant 1e-4 learning rate.
    \item NatPN: We first use $K$-means clustering with 64 clusters to obtain class labels $Y$ for the observations $X=O_t$. We then consider the case where $Y$ follows a categorical distribution with a Dirichlet prior on the distribution parameters. To lean the parameters, we then follow \citep{charpentier2022natural} to use the tabular encoder with 16 flow layers. We set the learning rate to be 1e-3 and train for a maximum of 1000 epochs.

    \item RND: On simulation, we use a 4x smaller network with identical architecture as the policy network, which takes in both $A_t$ and $O_t$ as inputs ($O_t$ as the conditioning variable). We train for 200 epochs with a batch size of 128, using the Adam optimizer with a constant 1e-4 learning rate. 
    On real data, we use network with the same size as the policy network to improve performance. We train for 2000 epochs with a batch size of 512, using the Adam optimizer with a constant 1e-4 learning rate. 
    During inference, a high $D_M(A_t,O_t;\hat{\theta})$ indicates a large mismatch between the predictor and target outputs, which we hypothesize results from the pair $(A_t,O_t)$ not being from a successful trajectory. 
\end{enumerate}

\begin{table}[!t]
    \caption{\small Hyperparams in evaluation protocol. We include the details for the policy networks in \cref{tab:params_policy} and the hyperparameters for CP band calibration in simulation in \cref{tab:params_CP_sim} and in robot hardware experiments in \cref{tab:params_CP_real_frt,tab:params_CP_real_cus}. For simulation tasks, training is done on one NVIDIA RTX A6000 GPU with 48GB memory. For the experiments on hardware, training is done on eight NVIDIA A100-SXM4-80GB GPUs with 80GB memory.}
    \label{tab:eval_protocol_params}
    \begin{minipage}{\linewidth}
        \centering
        \subcaption{Policy network}\label{tab:params_policy}
        \resizebox{\linewidth}{!}{
        \begin{tabular}{l|ccc}
        \hline
         & \makecell[c]{Dimension of\\ $(A_t, O_t)$} & \makecell[c]{Architecture of \\ (policy $g$, visual encoder for $O_t$)} & \makecell[c]{Policy training specification \\ (optimizer, lr, lr scheduler, batch size, number of epochs)} \\
         (Simulation) Square & (160, 274) & (UNet \citep{janner2022planning}, ResNet \citep{he2016deep}) & (AdamW \citep{loshchilov2018decoupled}, 1e-4, cosine \citep{loshchilov2017sgdr}, 64, 800) \\
         (Simulation) Can & (160, 274)& (UNet \citep{janner2022planning}, ResNet \citep{he2016deep}) & (AdamW \citep{loshchilov2018decoupled}, 1e-4, cosine \citep{loshchilov2017sgdr}, 64, 800) \\
         (Simulation) Toolhang & (160, 274)& (UNet \citep{janner2022planning}, ResNet \citep{he2016deep}) & (AdamW \citep{loshchilov2018decoupled}, 1e-4, cosine \citep{loshchilov2017sgdr}, 64, 300) \\
         (Simulation) Transport & (320, 548)& (UNet \citep{janner2022planning}, ResNet \citep{he2016deep}) & (AdamW \citep{loshchilov2018decoupled}, 1e-4, cosine \citep{loshchilov2017sgdr}, 64, 300) \\
         (Real) FoldRedTowel & (320, 4176)& (UNet \citep{janner2022planning}, ResNet \citep{he2016deep}) & (AdamW \citep{loshchilov2018decoupled}, 1e-4, cosine \citep{loshchilov2017sgdr}, 96, 1000) \\
         (Real) CleanUpSpill & (320, 6732)& (UNet \citep{janner2022planning}, CLIP \citep{radford2021learning})& (AdamW \citep{loshchilov2018decoupled}, 1e-4, cosine \citep{loshchilov2017sgdr}, 36, 500) \\
         \hline
        \end{tabular}}
        \vspace{10pt}
    \end{minipage}  
    \begin{minipage}{\linewidth}
        \centering
        \subcaption{(Simulation) CP band calibration and testing}\label{tab:params_CP_sim}
        \resizebox{\linewidth}{!}{
        \begin{tabular}{l|cccccccc}
        \hline
         & Square ID & Square OOD & Can ID & Can OOD & Toolhang ID & Toolhang OOD & Transport ID & Transport OOD \\
         \hline
         CP band modulation & \cref{eq:v2} & \cref{eq:v2} & \cref{eq:v2} & \cref{eq:v2} & \cref{eq:v2} & \cref{eq:v2} & \cref{eq:v2} & \cref{eq:v2} \\
         CP significance level & 0.05 & 0.05 & 0.05 & 0.05 & 0.05 & 0.05 & 0.05 & 0.05 \\ 
         \makecell[l]{(FM policy) Num successes  \\ for CP band mean} & 269 & 269 & 296 & 296 & 224 & 224 & 253 & 253  \\
         \makecell[l]{(FM policy) Num successes \\ for CP band width} & 629 & 629 & 693 & 693 & 523 & 523 & 591 & 591  \\
         \makecell[l]{(FM policy) Num test rollouts \\ for evaluation} & 1000 & 2000 & 1000 & 2000 & 1000 & 2000 & 1000 & 2000 \\
         \makecell[l]{(DP) Num successes  \\ for CP band mean} & 31 & 31 & 34 & 34 & 27 & 27 & 27 & 27  \\
         \makecell[l]{(DP) Num successes \\ for CP band width} & 82 & 82 & 90 & 90 & 70 & 70 & 71 & 71\\
         \makecell[l]{(DP) Num test rollouts \\ for evaluation} & 125 & 250 & 125 & 250& 125 & 250& 125 & 250 \\
         \hline
        \end{tabular}}
        \vspace{10pt}
    \end{minipage}    
    \begin{minipage}{\linewidth}
        \centering
        \setstretch{1.25}
        \subcaption{(Hardware: FoldRedTowel) CP band calibration and testing}\label{tab:params_CP_real_frt}
        \resizebox{0.65\linewidth}{!}{
        \begin{tabular}{l|cccc}
        \hline
         & \makecell[c]{ID Disturb \\ (Setting-dependent)} & \makecell[c]{ID Disturb \\ (ID-only)} & \makecell[c]{OOD \\ (Setting-dependent)} & \makecell[c]{OOD \\ (ID-only)} \\
         \hline
         CP band modulation & \cref{eq:v1} & \cref{eq:v1} & \cref{eq:v1} & \cref{eq:v1} \\
         CP significance level & 0.05 & 0.05 & 0.05 & 0.05 \\
         \makecell[l]{Num successes for \\  CP band mean} & 7 & 7 & 4 & 7 \\
         \makecell[l]{Num successes for \\ CP band width} & 23 & 23 & 13 & 23 \\
         \makecell[l]{Num test rollouts for \\ evaluation} & 20 & 50 & 20 & 50 \\
         \hline
        \end{tabular}}
        \vspace{10pt}
    \end{minipage}
    \begin{minipage}{\linewidth}
        \centering
        \setstretch{1.25}
        \subcaption{(Hardware: CleanUpSpill) CP band calibration and testing}\label{tab:params_CP_real_cus}
        \resizebox{0.65\linewidth}{!}{
        \begin{tabular}{l|cccc}
        \hline
         & \makecell[c]{OOD (DP) \\ (Setting-dependent)} & \makecell[c]{OOD (DP) \\ (ID-only)} & \makecell[c]{OOD (FM policy) \\ (Setting-dependent)} & \makecell[c]{OOD (FM policy) \\ (ID-only)} \\
         \hline
         CP band modulation & \cref{eq:v2} & \cref{eq:v2} & \cref{eq:v1} & \cref{eq:v1} \\
         CP significance level & 0.05 & 0.05 & 0.05 & 0.05 \\
         \makecell[l]{Num successes for \\  CP band mean} & 6 & 5 & 6 & 9 \\
         \makecell[l]{Num successes for \\ CP band width} & 14 & 12 & 14 & 18 \\
         \makecell[l]{Num test rollouts for \\ evaluation} & 20 & 50 & 20 & 50 \\
         \hline
        \end{tabular}}
    \end{minipage}
\end{table}
\begin{figure}[!t]
    \centering
    \begin{minipage}{0.24\linewidth}
        \includegraphics[width=\linewidth]{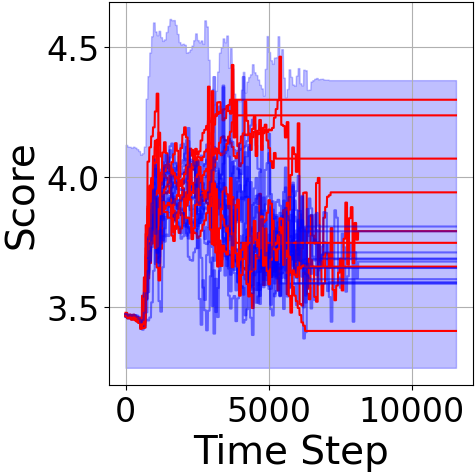}
        \subcaption{PCA-kmeans}
    \end{minipage}
    \begin{minipage}{0.24\linewidth}
        \includegraphics[width=\linewidth]{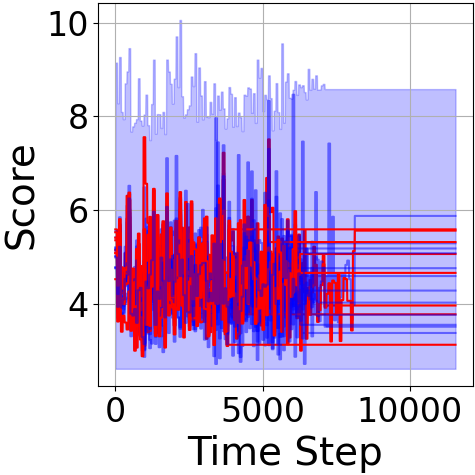}
        \subcaption{SPARC}
    \end{minipage}
    \begin{minipage}{0.24\linewidth}
        \includegraphics[width=\linewidth]{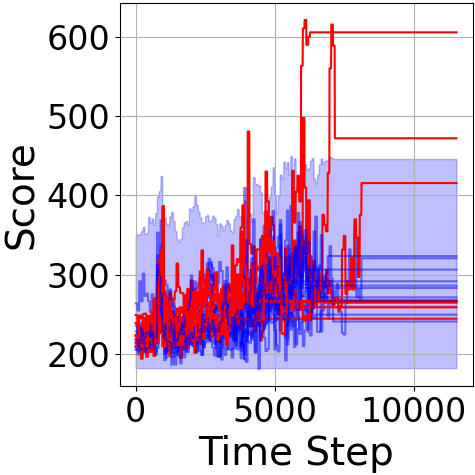}
        \subcaption{logpZO}
    \end{minipage}
    \begin{minipage}{0.24\linewidth}
        \includegraphics[width=\linewidth]{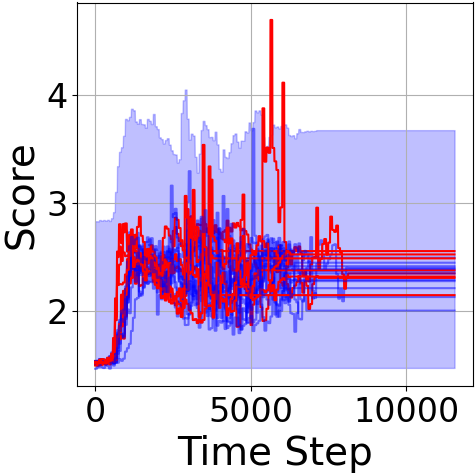}
        \subcaption{RND}
    \end{minipage}
    \caption{\small Qualitative results of detection scores overlaid with CP bands on the real \textbf{FoldRedTowel OOD} task. The layout is the same as \cref{fig:score_visual}. We notice that spikes of scores computed on failed trajectories are more evident for the learnt logpZO and RND than for the post-hoc PCA-kmeans and SPARC.}
    \label{fig:score_visual_real}
\end{figure}

\begin{figure}[!t]
    \centering
    \begin{minipage}{0.49\textwidth}
        \includegraphics[width=\linewidth]{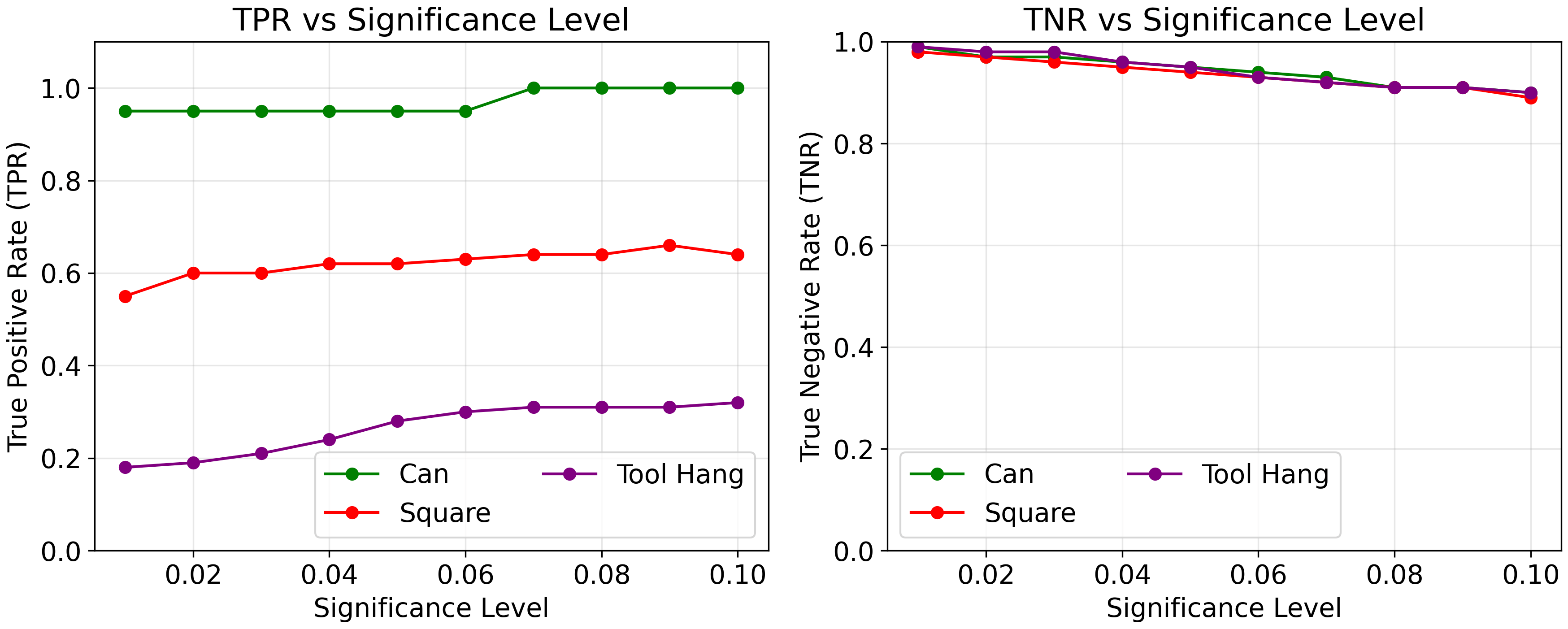}
        \subcaption{Simulation}
    \end{minipage}
    \begin{minipage}{0.49\textwidth}
        \includegraphics[width=\linewidth]{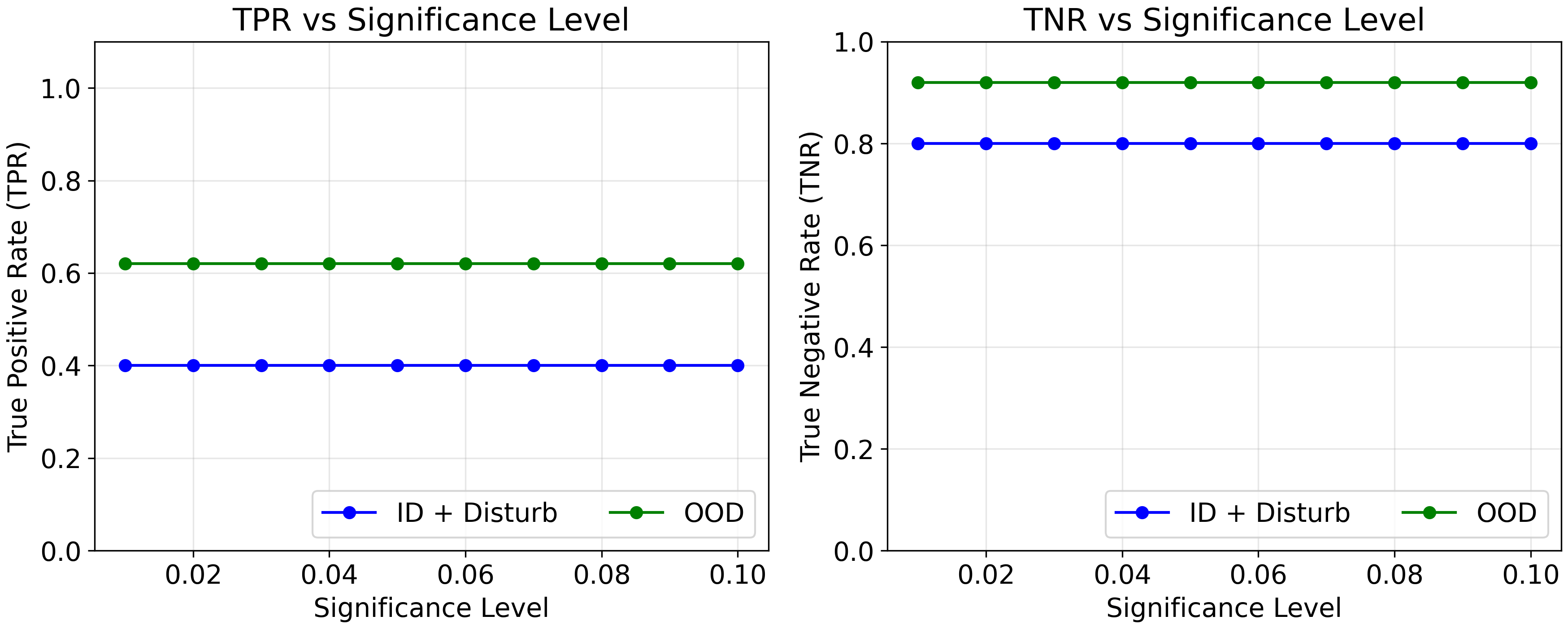}
        \subcaption{Robot hardware (\textbf{FoldRedTowel})}
    \end{minipage}
    \caption{TPR and TNR vs. CP significance level in simulation and hardware experiments.}
    \label{fig:alpha_ablate}
\end{figure}

\begin{figure}[!t]
    \centering
    \begin{minipage}[b]{\linewidth}
        \includegraphics[width=\linewidth]{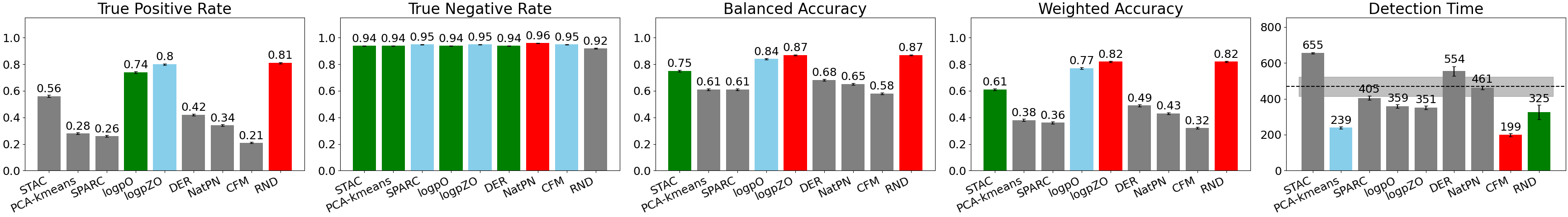}
        \vspace{-20pt}
        \subcaption{Transport ID}
    \end{minipage}
    \begin{minipage}[b]{\linewidth}
        \includegraphics[width=\linewidth]{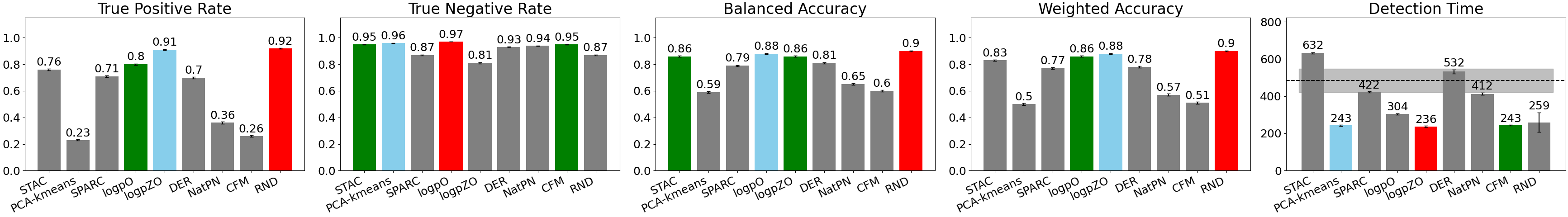}
        \vspace{-20pt}
        \subcaption{Transport OOD}
    \end{minipage}
    \begin{minipage}[b]{\linewidth}
        \includegraphics[width=\linewidth]{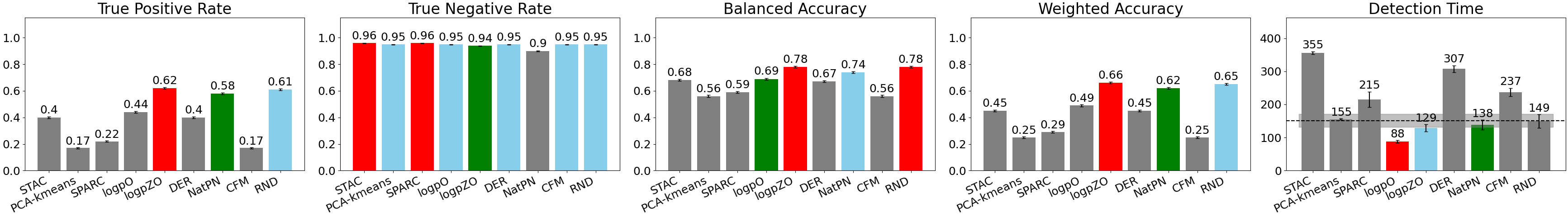}
        \vspace{-20pt}
        \subcaption{Square ID}
    \end{minipage}
    \begin{minipage}[b]{\linewidth}
        \includegraphics[width=\linewidth]{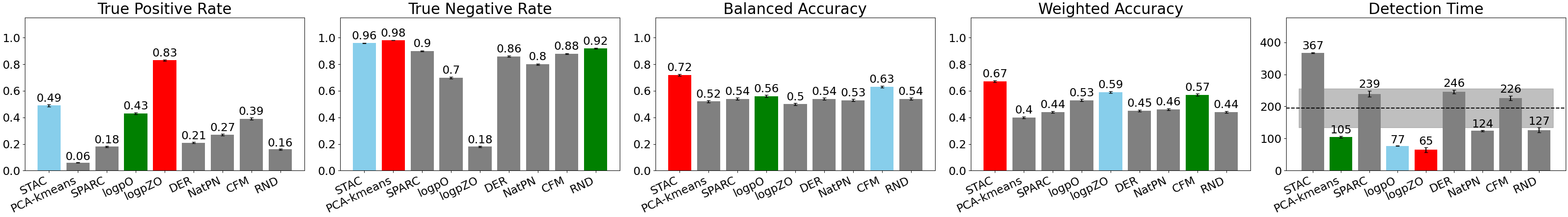}
        \vspace{-20pt}
        \subcaption{Square OOD}
    \end{minipage}
    \begin{minipage}[b]{\linewidth}
        \includegraphics[width=\linewidth]{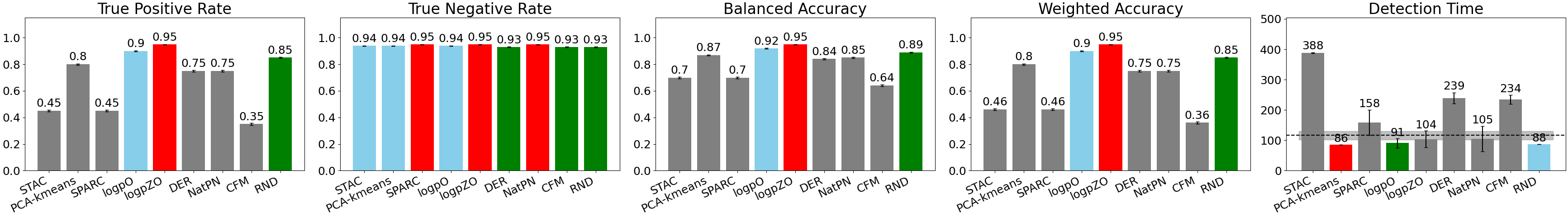}
        \vspace{-20pt}
        \subcaption{Can ID}
    \end{minipage}
    \begin{minipage}[b]{\linewidth}
        \includegraphics[width=\linewidth]{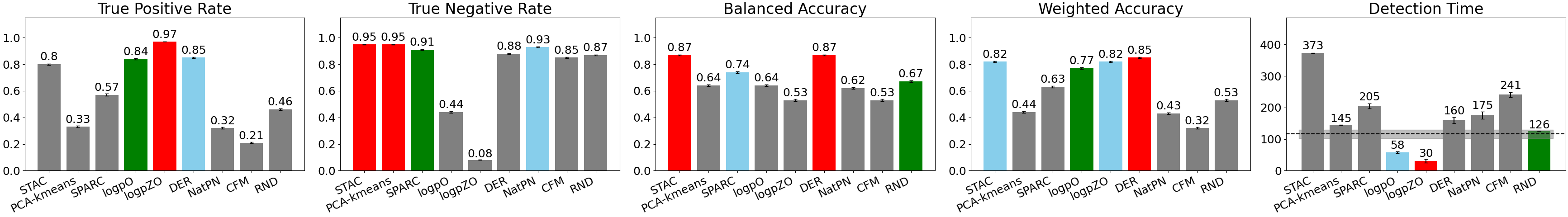}
        \vspace{-20pt}
        \subcaption{Can OOD}
    \end{minipage}
    \begin{minipage}[b]{\linewidth}
        \includegraphics[width=\linewidth]{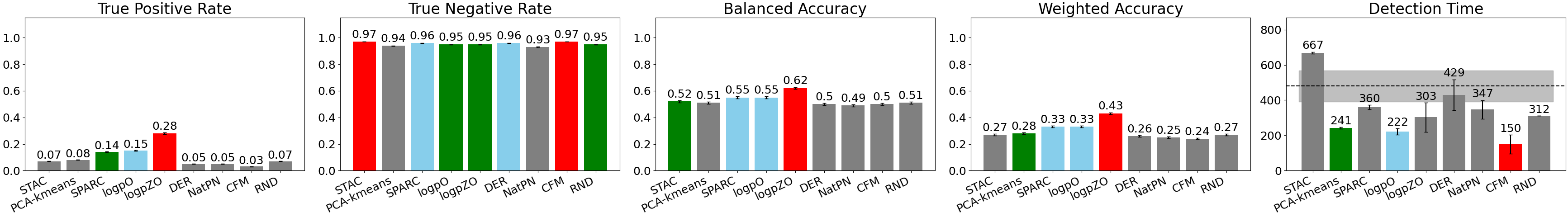}
        \vspace{-20pt}
        \subcaption{Toolhang ID}
    \end{minipage}
    \begin{minipage}[b]{\linewidth}
        \includegraphics[width=\linewidth]{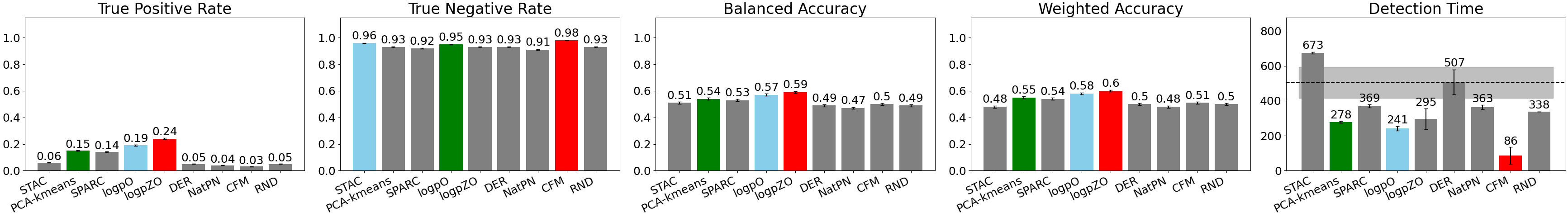}
        \vspace{-20pt}
        \subcaption{Toolhang OOD}
    \end{minipage}
    \caption{\small  Quantitative results in simulation tasks by FM policy ({\color{red} best}, {\color{cyan} second}, {\textcolor{ForestGreen}{third}}), which augments \cref{fig:sim_metrics_abbrev} by including all quantitative metrics. 
    The takeaways are similar as before, where logpZO and RND are the top-2 best-performing method overall.}
    \label{fig:sim_metrics}
\end{figure}

\begin{figure}[!t]
    \centering
    \begin{minipage}[b]{\linewidth}
        \includegraphics[width=\linewidth]{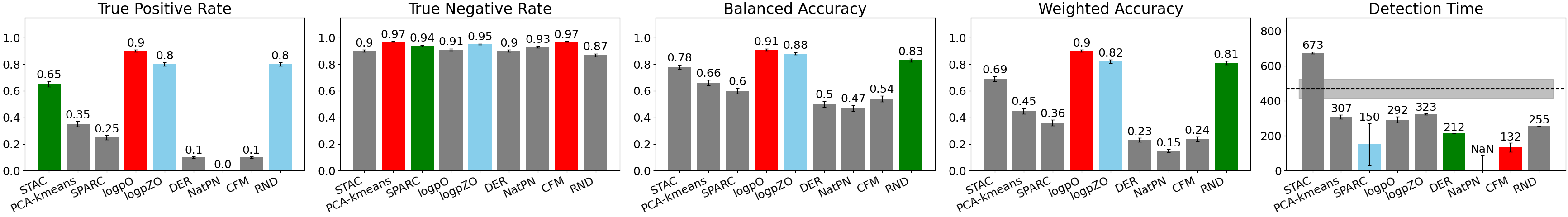}
        \vspace{-20pt}
        \subcaption{Transport ID}
    \end{minipage}
    \begin{minipage}[b]{\linewidth}
        \includegraphics[width=\linewidth]{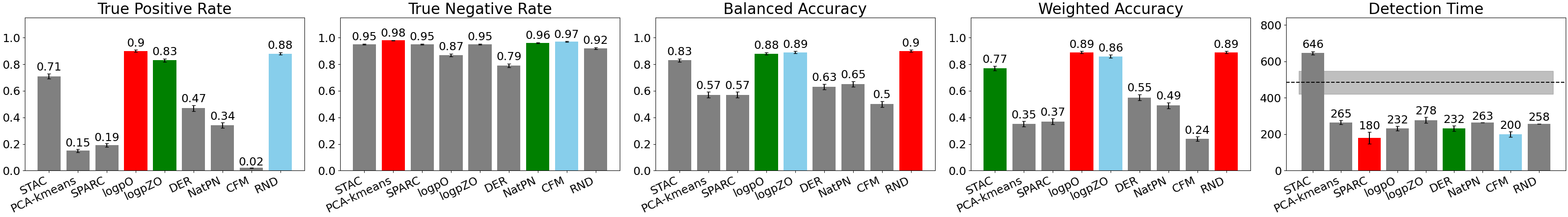}
        \vspace{-20pt}
        \subcaption{Transport OOD}
    \end{minipage}
    \begin{minipage}[b]{\linewidth}
        \includegraphics[width=\linewidth]{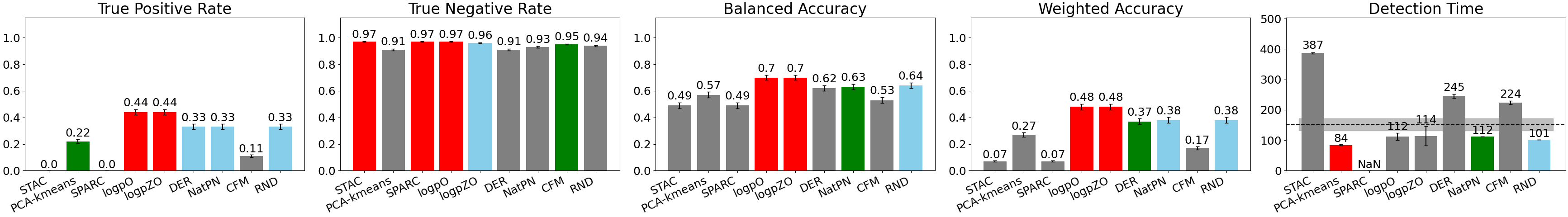}
        \vspace{-20pt}
        \subcaption{Square ID}
    \end{minipage}
    \begin{minipage}[b]{\linewidth}
        \includegraphics[width=\linewidth]{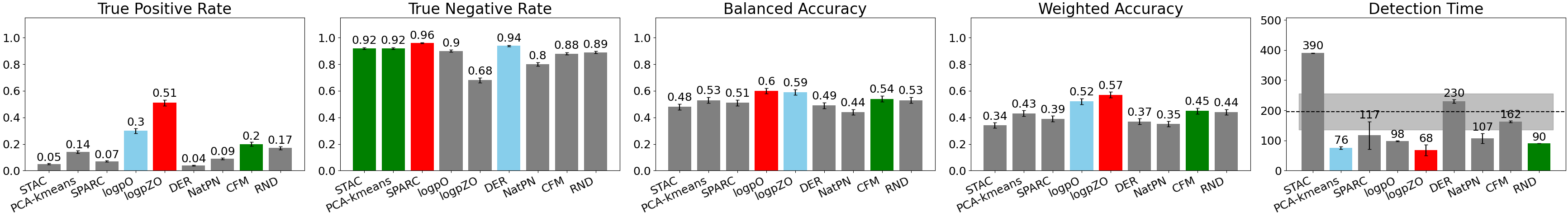}
        \vspace{-20pt}
        \subcaption{Square OOD}
    \end{minipage}
    \begin{minipage}[b]{\linewidth}
        \includegraphics[width=\linewidth]{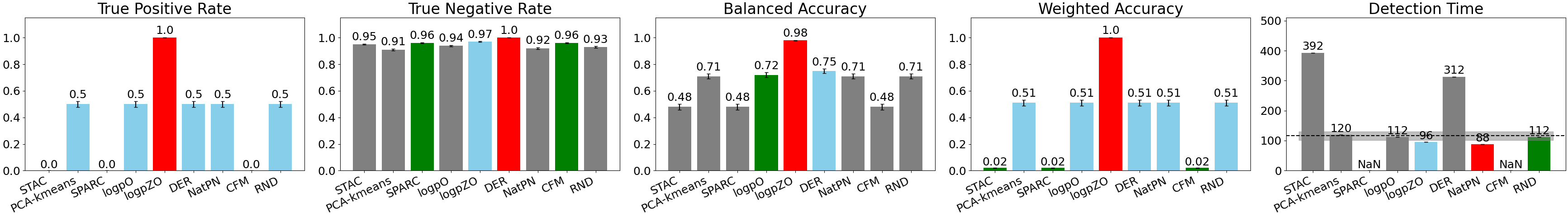}
        \vspace{-20pt}
        \subcaption{Can ID}
    \end{minipage}
    \begin{minipage}[b]{\linewidth}
        \includegraphics[width=\linewidth]{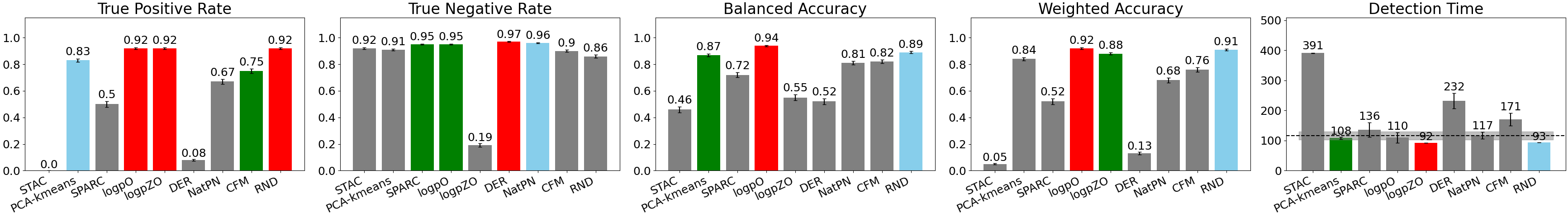}
        \vspace{-20pt}
        \subcaption{Can OOD}
    \end{minipage}
    \begin{minipage}[b]{\linewidth}
        \includegraphics[width=\linewidth]{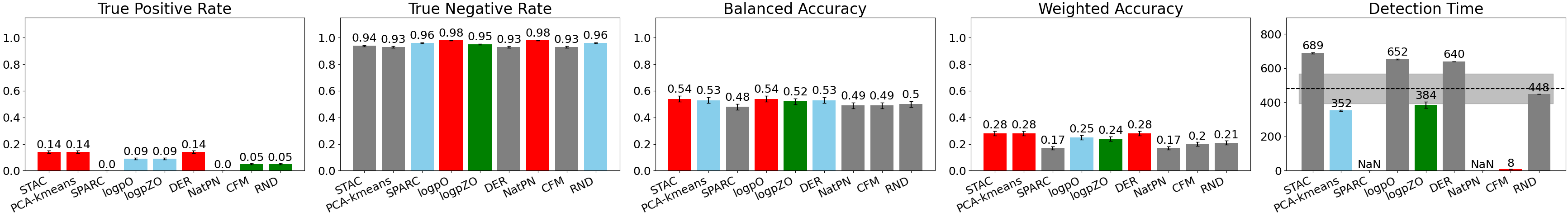}
        \vspace{-20pt}
        \subcaption{Toolhang ID}
    \end{minipage}
    \begin{minipage}[b]{\linewidth}
        \includegraphics[width=\linewidth]{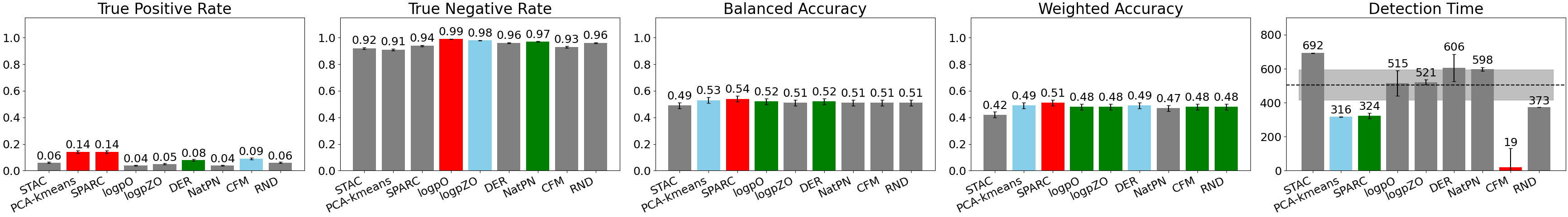}
        \vspace{-20pt}
        \subcaption{Toolhang OOD}
    \end{minipage}
    \caption{\small  Quantitative results in simulation tasks by DP ({\color{red} best}, {\color{cyan} second}, {\textcolor{ForestGreen}{third}}). The layout is identical to \cref{fig:sim_metrics_abbrev}.
    We similarly observe that learned methods seem to have more capacity to detect failures than post-hoc ones, with logpZO and RND being the best-performing methods.}
    \label{fig:sim_metric_DP}
\end{figure}

\begin{figure}[!t]
    \centering
    \begin{minipage}[b]{\linewidth}
        \includegraphics[width=\linewidth]{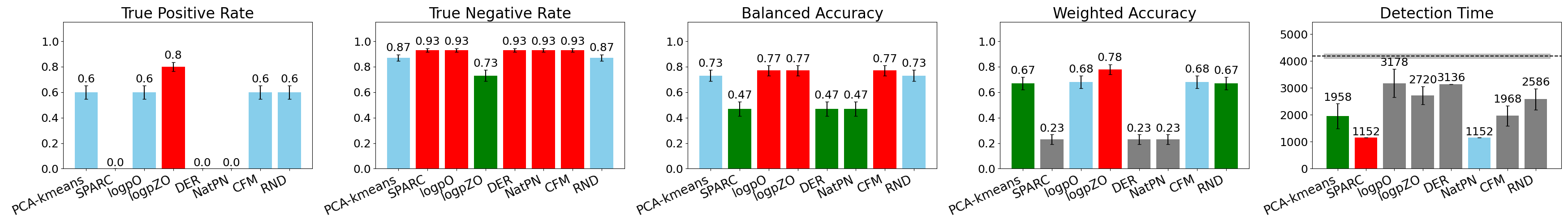}
        \subcaption{FM policy: (Setting-dependent band) ID + Disturb}
    \end{minipage}
    \begin{minipage}[b]{\linewidth}
        \includegraphics[width=\linewidth]{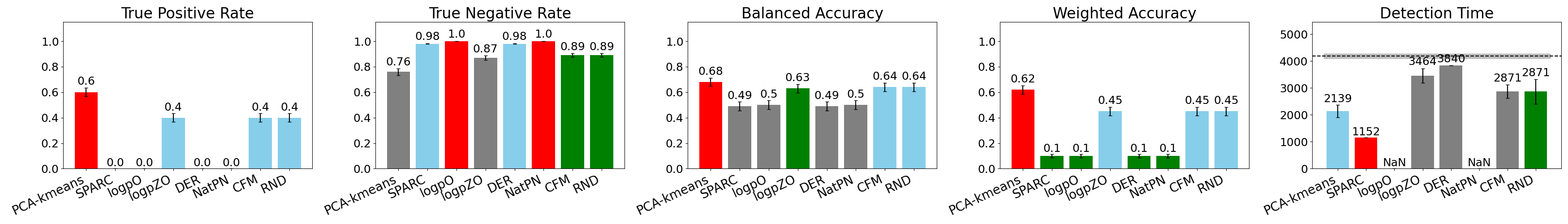}
        \subcaption{FM policy: (ID-only band) ID + Disturb}
    \end{minipage}
    \begin{minipage}[b]{\linewidth}
        \includegraphics[width=\linewidth]{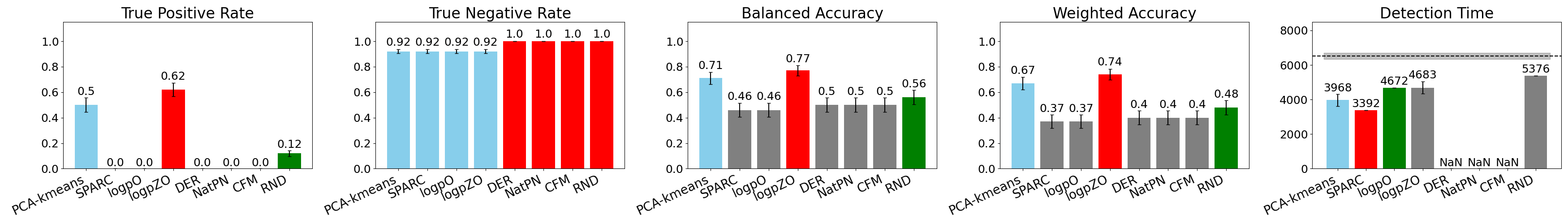}
        \subcaption{FM policy: (Setting-dependent band) OOD initial condition}
    \end{minipage}
    \begin{minipage}[b]{\linewidth}
        \includegraphics[width=\linewidth]{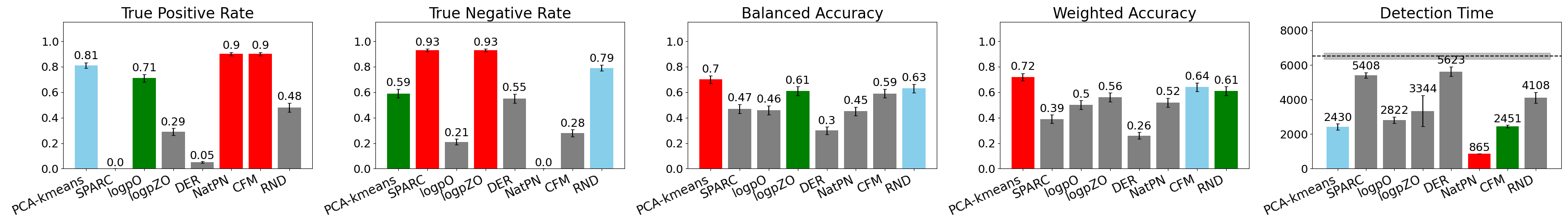}
        \subcaption{FM policy: (ID-only band) OOD initial condition}
    \end{minipage}
    \caption{\small  Quantitative results on the \textbf{FoldRedTowel} robot hardware task using two ways to compute the CP band ({\color{red} best}, {\color{cyan} second}, {\textcolor{ForestGreen}{third}}).
    logpZO remains to be the most robost method overall.}
    \label{fig:real_metrics}
\end{figure}

\begin{figure}[!t]
    \centering
    \begin{minipage}[b]{\linewidth}
        \includegraphics[width=\linewidth]{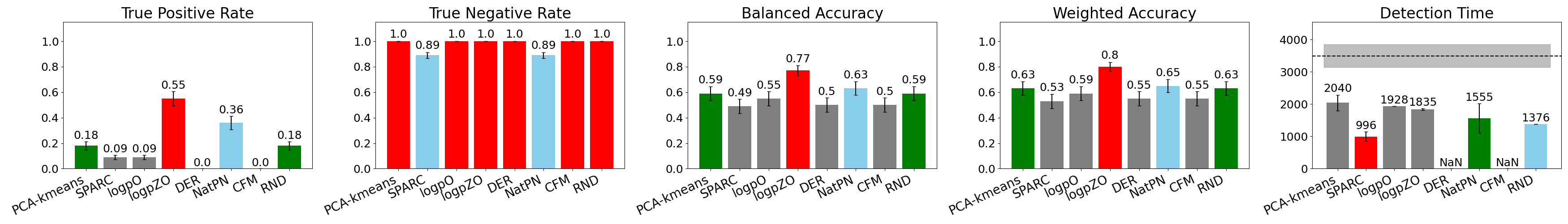}
        \subcaption{DP: (Setting-dependent band) OOD initial condition}
    \end{minipage}
    \begin{minipage}[b]{\linewidth}
        \includegraphics[width=\linewidth]{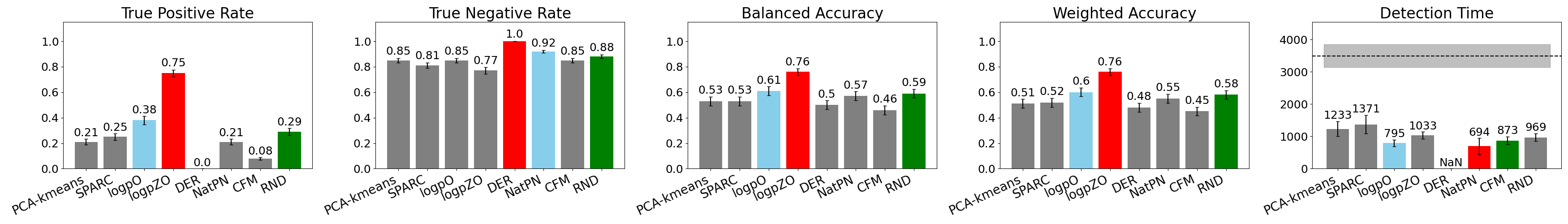}
        \subcaption{DP: (ID-only band) OOD initial condition}
    \end{minipage}
    \begin{minipage}[b]{\linewidth}
        \includegraphics[width=\linewidth]{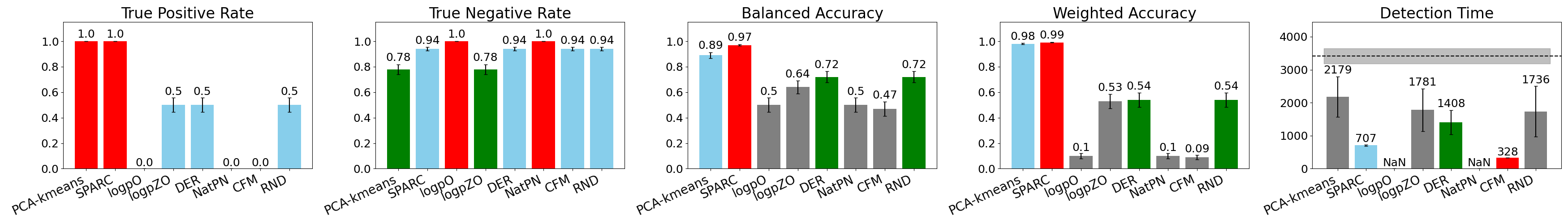}
        \subcaption{FM policy: (Setting-dependent band) OOD initial condition}
    \end{minipage}
    \begin{minipage}[b]{\linewidth}
        \includegraphics[width=\linewidth]{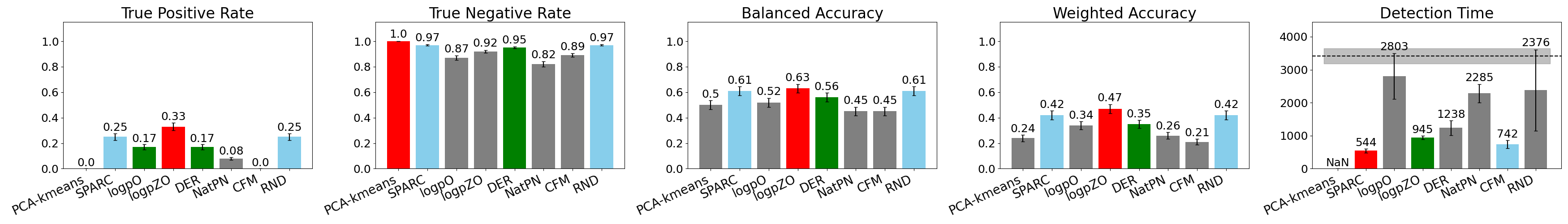}
        \subcaption{FM policy: (ID-only band) OOD initial condition}
    \end{minipage}
    \caption{\small  Quantitative results on the \textbf{CleanUpSpill} robot hardware task using two ways to compute the CP band ({\color{red} best}, {\color{cyan} second}, {\textcolor{ForestGreen}{third}}).
    logpZO remains to be the most robost method overall.}
    \label{fig:real_metrics_Spill}
\end{figure}

\subsection{Ablation}\label{sec:ablation}
We conduct ablation studies on 
the behavior of our method under varying CP significance levels $\alpha$.


In \cref{fig:alpha_ablate}, we show TPR and TNR for 10 equally spaced values of $\alpha \in [0.01, 0.1]$ using logpZO. As expected, higher $\alpha$ increases TPR and decreases TNR, since more rollouts are flagged as failures. This trend is clearer in simulation; in hardware experiments, the effect is muted due to the limited number of rollouts and therefore constant calibration quantiles for small $\alpha$. Overall, $\alpha=0.05$ offers a robust trade-off, which is what we used in all experiments.


\end{document}